%% file: main.tex
\documentclass{article}

\usepackage[preprint]{neurips_2022}

\usepackage[utf8]{inputenc} % allow utf-8 input
\usepackage[T1]{fontenc}    % use 8-bit T1 fonts
\usepackage{hyperref}       % hyperlinks
\hypersetup{colorlinks,allcolors=[rgb]{0,0.07843,0.45098}}
\usepackage{url}            % simple URL typesetting
\usepackage{booktabs}       % professional-quality tables
\usepackage{amsfonts}       % blackboard math symbols
\usepackage{nicefrac}       % compact symbols for 1/2, etc.
\usepackage{microtype}      % microtypography

\usepackage{listings}
\usepackage{xcolor}
\usepackage[normalem]{ulem}
\usepackage{soul}
\newcommand{\greenuline}[1]{{\color{LimeGreen}\uline{{\color{RoyalBlue}#1}}}}
\newcommand{\purpleuline}[1]{{\color{Purple}\uline{{\color{RoyalBlue}#1}}}}

\definecolor{Cerulean}{rgb}{0,0,0.95}
\definecolor{LimeGreen}{rgb}{0.15,0.65,0.15}
\definecolor{RoyalBlue}{rgb}{0.25,0.41,0.88}
\definecolor{Rose}{rgb}{1.0, 0.15, 0.21}
\definecolor{Orange}{rgb}{1.0, 0.5, 0.0}
\definecolor{Gray}{gray}{0.6}
\definecolor{Black}{gray}{0.0}
\definecolor{Purple}{rgb}{0.77,0.12,0.64}
\definecolor{codegreen}{rgb}{0,0.8,0}
\definecolor{codered}{rgb}{0.95,0,0.3}
\definecolor{codegray}{rgb}{0.5,0.5,0.5}
\definecolor{codepurple}{rgb}{0.58,0,0.82}
\definecolor{backcolour}{rgb}{0.95,0.95,0.95}
 
\lstdefinestyle{Python}{
    language        = Python,
    basicstyle      = \scriptsize\ttfamily,
    keywordstyle    = \color{black},
    keywordstyle    = [2] \color{black}, % just to check that it works
    stringstyle     = \color{black},
    commentstyle    = \color{blue}\ttfamily,
    backgroundcolor = \color{backcolour},
    breakatwhitespace=false,
    breaklines=true,
    basewidth=0.55em,
    tabsize=2
}

\usepackage{float}
\usepackage{inconsolata}
\usepackage{xcolor}  % Text colors
\usepackage{caption}
\usepackage{subcaption}
\usepackage{graphicx}
\usepackage{enumitem}

\usepackage{physics}
\usepackage{amsmath}
\usepackage{tikz}
\usepackage{mathdots}
\usepackage{yhmath}
\usepackage{cancel}
\usepackage{color}
\usepackage{siunitx}
\usepackage{array}
\usepackage{multirow}
\usepackage{amssymb}
\usepackage{gensymb}
\usepackage{tabularx}
\usepackage{extarrows}
\usepackage{booktabs}
\usepackage{wrapfig}
\usetikzlibrary{fadings}
\usetikzlibrary{patterns}
\usetikzlibrary{shadows.blur}
\usetikzlibrary{shapes}

\newcommand{\ie}{i.e., }
\newcommand{\eg}{e.g., }

\title{Socratic Models: Composing Zero-Shot \\ Multimodal Reasoning with Language}

\author{%
Andy Zeng, Maria Attarian, Brian Ichter, Krzysztof Choromanski, Adrian Wong, \\
\textbf{Stefan Welker, Federico Tombari, Aveek Purohit, Michael S. Ryoo,} \\
\textbf{Vikas Sindhwani, Johnny Lee, Vincent Vanhoucke, Pete Florence} \\\\
Google
\vspace{-1.0em}
}

\begin{document}

\maketitle

\begin{abstract}
% pete: foundation -> pretrained
Large pretrained %language-based
(\eg ``foundation'') models exhibit distinct capabilities depending on the domain of data they are trained on. While these domains are generic, they may only barely overlap.
For example, visual-language models (VLMs) are trained on Internet-scale image captions, 
but large language models (LMs) are further trained on Internet-scale text with no images (\eg spreadsheets, SAT questions, code).
As a result, these models store different forms of commonsense knowledge across different domains.
In this work, we show that this diversity is symbiotic, and can be leveraged through Socratic Models (SMs): a modular framework in which multiple pretrained
% foundation
models may be composed zero-shot \ie via multimodal-informed prompting,
to exchange information with each other and capture new multimodal capabilities, without requiring finetuning.
% through a guided language-based exchange
% language is the shared/common representation by which these models interact.
% with structured Socratic dialogue -- in which new multimodal tasks are formulated as a guided language-based exchange between different pre-existing foundation models, without additional finetuning.
% We present a case study of one such framework, Socratic Models (SMs), which
% To this end, we present a case study of
% To this end, we present a case study of Socratic Models (SMs), a framework that demonstrates how to connect language-interactable models for new tasks, without domain-specific data collection.
With minimal engineering, SMs are not only competitive with state-of-the-art zero-shot image captioning and video-to-text retrieval, %  with 44.7\% R@1 on MSR-VTT
but also enable new applications such as (i) answering free-form questions about egocentric video, (ii) engaging in multimodal assistive dialogue with people (\eg for cooking recipes) by interfacing with external APIs and databases (\eg web search), and (iii) robot perception and planning.
% In the context of egocentric perception, we present a case study of Socratic Models (SMs) that can provide meaningful results for complex tasks such as generating free-form answers to contextual questions about egocentric video, by formulating video Q\&A as short story Q\&A, \ie summarizing the video into a short story, then answering questions about it.
% Additionally, SMs can generate captions for Internet images, and are competitive with state-of-the-art on zero-shot video-to-text retrieval with 42.8 R@1 on MSR-VTT 1k-A.
% SMs demonstrate how to compose language-interactable models zero-shot to capture new multimodal functionalities, without domain-specific data collection. % Prototypes are available at \mbox{\href{https://socraticmodels.github.io/}{socraticmodels.github.io}}.
% had: %shed light on promising new opportunities to build simple systems that
\end{abstract}

\section{Introduction}\label{sec:intro}

Large pretrained models (\eg BERT \cite{devlin2018bert}, GPT-3 \cite{brown2020language}, CLIP \cite{radford2021learning}) have enabled impressive capabilities \cite{bommasani2021opportunities}: from zero-shot image classification \cite{radford2021learning,li2021align}, to high-level planning \cite{huang2022language,saycan2022arxiv}. Their capabilities depend on their training data -- while they
may be broadly
%generic or indiscriminately
crawled from the web, their distributions remain distinct across domains. For example, 
in terms of linguistic data, visual-language models (VLMs) \cite{wang2021simvlm,jain2021mural} are trained on image and video captions, but large language models (LMs) \cite{devlin2018bert,thoppilan2022lamda,chen2021evaluating} are additionally trained on a large corpora of other data such as spreadsheets, fictional novels, and standardized test questions.
These different domains offer distinct commonsense knowledge: VLMs can ground text to visual content, but LMs can perform a variety of other linguistic tasks (\eg reading comprehension %questions
\cite{rajpurkar2018know}). % that VLMs may struggle to address alone. %to date have not been demonstrated with VLMs alone.
%\vikas{I don't know how prevalent the term "foundation models" is particularly outside of NLP yet - may help to remind the reader by mentioning GPT3, Bert etc as prominent examples and the massive data/parameter scale underlying them.}
%pete: added specific callouts for those.
% In this work, we propose that these model differences are complementary \pete{There's maybe something that could be stronger here.  I think it's maybe not as surprising that they are complementary... not only are they complementary, but their complementary differences can be captured togehter zero-shot?  We might also consider a re-order in which the "Rather than scaling all the multimodal" comment comes before we announce what SMs are... which is composing the models, zero-shot.}, and can be jointly leveraged to build AI systems with structured Socratic dialogue -- in which new multimodal tasks are formulated as a guided exchange between different pre-existing language-interactable large pretrained (``foundation'') models, without additional finetuning.
% jointly leveraged through zero-shot composition
In this work, we propose these model differences are complementary and can be jointly leveraged to compose (via prompting) new multimodal capabilities out-of-the-box. % that the models may otherwise independently struggle to do.
To this end, we present Socratic Models\footnote{\small The name draws from an analogy to the Socratic Method, but between modules interacting through language.} (SMs), a modular framework in which new tasks are formulated as a language-based exchange between pretrained models and other modules, without additional training or finetuning. These modules can either contain (i) large pretrained (``foundation'' \cite{bommasani2021opportunities}) models, or (ii) APIs that interface with external capabilities or databases (\eg web search, robot actions).
Rather than scaling task-specific multimodal training data in the areas of overlap (\eg alt-text captions \cite{jia2021scaling}), or unifying model architectures for multitask learning \cite{hu2021transformer}, SMs embrace the zero-shot capabilities of pretrained models by prompt engineering guided multimodal discussions between the independent models to perform joint inference on a task-specific output.  % SMs use language as the representation
%common denominator (\ie semantically compressed and interpretable intermediate representation)
% by which inter-domain foundation models can jointly be used for inference.
%interact with each other to to form joint predictions.
% \vikas{Consider title simplification, \eg Zero-shot Ego-centric Visuo-Linguistic Reasoning via Multi-model Socratic Dialog?}
% \vincent{Vikas, your notion of 'simplification' defies my understanding :D}
% \pete{:)}

% augment them with access to external sources of knowledge
% predominantly explore modules that are 

\begin{wrapfigure}{r}{0.5\textwidth}
  %\vspace{-1.2em}
  \captionsetup{type=figure}
  \includegraphics[width=\linewidth]{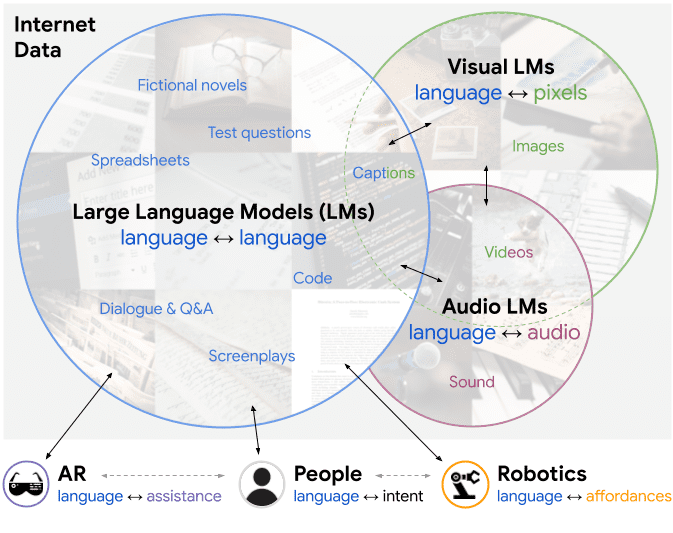}
  \vspace{-1.8em}
  \captionof{figure}{\small Large pretrained ``foundation'' models trained across different domains learn complementary forms of commonsense, and language is an intermediate representation by which these models can communicate with each other to generate joint predictions for new multimodal tasks, without requiring finetuning. New applications (\eg augmented reality (AR), human feedback, robotics) can be viewed as adding participants to the multi-model discussion. %In this paper, we study inter-model dialogue in the context of multimodal reasoning.
  % \pete{Add AR in figure} done
  % \vikas{I did not understand the reference to non-Bayesian above. Best avoided here.} done
  % https://docs.google.com/drawings/d/1MfDANylm8jYl_ZwFhfJ4Ec7RBqKBoiK2yCOKraxZ2tM/edit?usp=sharing
  }
  \label{fig:internet-venn}
  \vspace{-1.5em}
\end{wrapfigure}

\begin{wrapfigure}{r}{0.6\textwidth}
  \vspace{-1.1em}
  %\captionsetup{type=figure}
  \begin{subfigure}{0.6\textwidth}
  \centering
  \includegraphics[width=\linewidth]{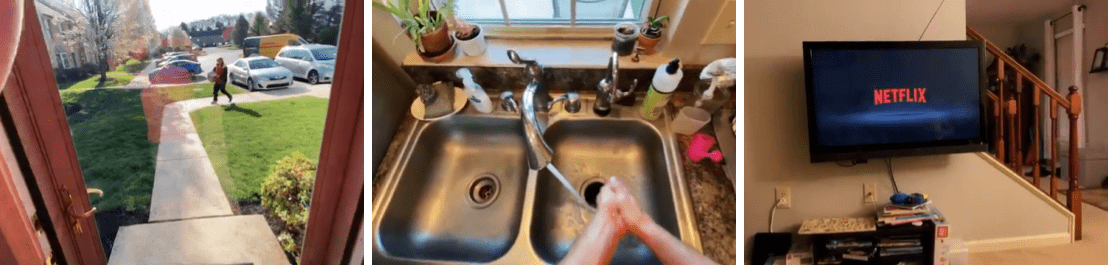}
  \end{subfigure}%
  \vspace{0.1em}
  \begin{subfigure}{0.6\textwidth}
  \centering
  \noindent\fcolorbox{lightgray}{white}{%
    \parbox{0.97\linewidth}{%
      \vspace{-0.5em}
      \begin{flushleft}\scriptsize{
        \texttt{%{\color{Black}Language-Based World-State History from Egocentric Video:\\~\\}
        {\color{Gray}01:45 PM: Places: {\color{LimeGreen}porch}. Objects: {\color{LimeGreen}package, porch, door}. Activities: {\greenuline{receiving}}. {\color{RoyalBlue}I was receiving a package}.\\
03:24 PM: Places: {\color{LimeGreen}kitchen}. Objects: {\color{LimeGreen}human hand, sink, human arm}. Activities: {\greenuline{washing dishes}}. {\color{RoyalBlue}I was washing dishes in a kitchen}.\\
07:20 PM: Places: {\color{LimeGreen}living room}. Objects: {\color{LimeGreen}netflix, television, shelf}. Activities: {\greenuline{watching netflix}}. {\color{RoyalBlue}I was watching netflix}.\\~\\
Question: {\color{magenta}When did I last wash my hands?}\\
Long answer: {\color{RoyalBlue}I last washed my hands at 3:24 PM}.\\
This is because {\color{RoyalBlue}I was washing dishes in a kitchen}.}}}
        \vspace{-0.5em}
      \end{flushleft}
    }%
  }
  \end{subfigure}%
  \vspace{-0.2em}
  \caption{\small SMs with {\color{LimeGreen}VLM}, {\color{RoyalBlue}LM}, and {\color{Purple}ALM} can be prompted to generate a captions for key moments in videos, which can be assembled into a language-based world-state history (\eg in the form of an event log) that the {\color{RoyalBlue}LM} can answer free-form questions about.}
  \vspace{-1.0em}
  \label{fig:teaser-ego}
%   \\~\\
% Q: Did I drive today?\\
% A: {\color{RoyalBlue}No, I did not drive today}. E: {\color{RoyalBlue}I was at home all day}.\\
% Q: {\color{magenta}Why did I go to the front porch today?}\\
% A: {\color{RoyalBlue}I went to the front porch today to receive a package}.\\
% Q: {\color{magenta}Why were my wife and I in the living room this evening?}\\
% A: {\color{RoyalBlue}My wife and I were in the living room this evening because we were watching netflix}.
\end{wrapfigure}

Across a number of domains spanning vision, language, and audio modalities -- %we find that specific instantiations of SMs with scripted policies that guide a closed-loop exchange between LMs, VLMs, and audio-language models (ALMs), and can generate meaningful results for a variety of tasks. % (Fig.~\ref{fig:teaser}).
and via a small amount of creative prompt-enabled multimodal composition -- SMs are quantitatively competitive with zero-shot state-of-the-art on standard benchmarks including (i) image captioning on MS COCO
\cite{chen2015microsoft, lin2014microsoft}, (ii) contextual image captioning and description (improving $11.3$ to $38.9$ captioning CIDEr on Concadia \cite{kreiss2021concadia}), and (iii) video understanding with video-to-text retrieval
(from $40.7$ to $44.7$ zero-shot R@1 on MSR-VTT \cite{xu2016msr}).
% SMs expand the set of capabilities
SMs also enable new capabilities across applications such as (i) open-ended reasoning for egocentric perception (Fig.~\ref{fig:teaser-ego}), (ii) multimodal assistive dialogue to guide a user through a cooking recipe, and (iii) robot perception-driven planning for sequential pick and place.
% We present results on Internet image captioning (Sec.~\ref{sec:internet-image}) and the common video understanding task of video-to-text retrieval (Sec.~\ref{sec:internet-video-to-text}), but our highlighted application is .
SMs give rise to new opportunities to address classically challenging problems in one domain, by reformulating it as a problem in another.
For example, answering free-form questions about first-person videos (\eg {\em{“why did I go to the front porch today?”}}) was previously thought to be out-of-reach for egocentric perception without domain-specific data collection \cite{grauman2021ego4d,damen2020rescaling}. 
We show that this is possible with SMs by assembling video into a {\em{language-based world-state history}} (in the form of a short story, or event log), then performing various types of open-ended text-prompted tasks (\eg answering questions) about that world-state history -- \ie formulating video understanding as a reading comprehension problem, for which modern LMs are proficient.

The goal of this paper is (1) to discuss new perspectives on building AI systems that embrace the heterogeneity of pretrained models through structured Socratic dialogue, and (2) to give example demonstrations of what is already possible today with SMs on challenging multimodal tasks.  Our primary contribution is (i) the Socratic Models framework, which proposes to compose multimodal pretrained models through language, without requiring training.
% To the best of our knowledge, this proposition has not been studied in prior work.
The SMs framework contains key, enabling components such as the demonstrated (ii) multimodal prompting methods, including (iii) language-based world-state history for video understanding. Additional contributions include (iv) demonstrating strong quantitative performance of example SM systems, setting new zero-shot state-of-the-art on multiple tasks, including in image captioning %(\ie, $0.11 \rightarrow 0.38$ CIDEr on Concadia)
and video understanding, and (v) providing additional application examples on open-ended egocentric perception, multimodal assistants, and robot perception and planning. 
Our demonstrated SM systems are not without limitations -- we discuss the
% We also discuss the limitations of SM systems, such as
unreliability inherited from the models on which they are constructed, together with other potential broader impacts (Sec.~\ref{sec:discussion}). Code is available at \mbox{\href{https://socraticmodels.github.io/}{socraticmodels.github.io}}.

\section{Problem Setting, Background, and Related Work}\label{sec:background-and-related} % Problem Setting, and Preliminaries

{\textbf{Problem setting.}} We are interested in creating a variety of multimodal \cite{ngiam2011multimodal} applications enabled by large pretrained models, which can be viewed as a form of transfer \cite{caruana1997multitask,thrun1998lifelong}: ``knowledge'' learned from a set of {\em{surrogate tasks}} (\eg text completion, image-text similarity) is applied to new downstream {\em{target tasks}} (\eg image captioning, robot planning). Consider a set of target tasks where % $\mathbb{T} = \{T^i\}$, where each $T^i$ 
each task $i$ seeks a desired map $f^{i}: \mathcal{X}^i \rightarrow \mathcal{Y}^i$. %between an input space $\mathcal{X}^i$ and some output space $\mathcal{Y}^i$. 
We are particularly interested in cases where: (i) each input $\mathcal{X}^i$ and/or output $\mathcal{Y}^i$ may contain multiple modalities \eg from the power set of $\{${\color{RoyalBlue}language}, {\color{LimeGreen}vision}, {\color{Purple}audio}, {\color{Orange}robot actions}$\}$; (ii) there may be many such tasks; % \ie $|\mathbb{T}|$ is large;
(iii) each target task may have little or no training data available; and
(iv) models pretrained on the surrogate tasks are available. % $\{\mathcal{T}^j\}$. % $\mathbb{T}' = \{T'^j\}$;

\textbf{Pretraining weights} is a dominant paradigm for transfer learning with deep models, in which pretrained model weights (from surrogate tasks) 
% \textit{The ``pretrained Weights'' Approach.}
% To benefit from  pretrained models, the common machine learning paradigm can be described as the ``pretrained weights'' approach, in which parameters from models pretrained on tasks $\mathbb{T}'$ %$\bar{\theta}' \subset \bigcup \{\theta'\}$
are used to initialize some subset of parameters in the model for the target task,
which are then either (a) left frozen, or (b) finetuned.  %In this approach, for a new task $T$ a modeler typically determines how to leverage pretrained models $\{f'_{\theta'}\}$ by using the parameters $\{\theta'\}$ of those models.  
Pretraining deep models has been studied extensively in the unsupervised setting \cite{hinton2006fast,bengio2006greedy,vincent2008extracting,raina2007self,mesnil2012unsupervised}, and in the supervised setting was perhaps most popularized by ImageNet \cite{deng2009imagenet} pretraining \cite{girshick2014rich,donahue2014decaf,zeiler2014visualizing,sermanet2013overfeat}, Various forms of pretraining have been ubiquitous in NLP \cite{mikolov2013distributed,pennington2014glove,dai2015semi,ramachandran2016unsupervised,petersdeep,devlin2018bert,brown2020language}.  For each target task, model architectures and/or training procedures may need to be developed that are composed of these pretrained parameters, for which % that well leverage the pretraining.
domain expertise may be advantageous.  In multimodal training, it is common to leave sub-portions of models, for example ones associated with one but not other modalities, frozen for downstream tasks \cite{zhai2021lit,kulkarni2019unsupervised,florence2019self,tsimpoukelli2021multimodal,zakka2022xirl}.
%The approach has been highly successful in increasing performance on well-studied public benchmarks, and has been successful in transferring knowledge from one domain A to a domain B with some shared, but not completely shared, task characteristics -- for example, ImageNet-pretrained weights have been shown to, in some cases, improve real-world visuomotor learning for robots (cite).

{\textbf{Joint training of all modalities}} on specific target tasks is a common approach to multimodal learning \cite{tsimpoukelli2021multimodal,lu2019vilbert,mokady2021clipcap,gao2021clip2tv,song2022clip,zellers2022merlot}.
For each task $i$ one may
obtain a large multimodal dataset and train a task-specific map $f^i_{\theta_i}$ with parameters $\theta_i$, some of which may come from pretrained weights, either frozen or finetuned.
%$\{\mathbf{x}^i_j, \mathbf{y}^i_j\}_{j=1}^N$ of $N$ supervised samples and to optimize a model $f^i_{\theta}$, with parameters $\theta$, under some loss $\mathcal{L}$. This can be addressed by solving the optimization $\theta^* = \text{argmin}_{\theta} \  \mathcal{L}(\{\mathbf{x}_i, \mathbf{y}_i\}_{i=1}^N)$
%The task $T$ may differ from each task in $\{T'\}$ either in the input/output spaces $\mathcal{X}$ and $\mathcal{Y}$, or may share the same $\mathcal{X}$ and $\mathcal{Y}$ but be defined over a different distribution.
%We are also particularly interested in cases where (ii) $T$ may inherently involve multiple modalities (\eg language, vision, audio), and (iii) $T$ may have few or no training examples available, \ie $N$ is 0 or small.
A benefit of this approach is that it follows the playbook of: (1) curate a big dataset, (2) train a big model, which given enough data and compute has proven to be formidable \cite{sutskever2014sequence}.

Combining both weights from large pretrained models with multimodal joint training, several works have achieved strong results for a number of downstream multimodal applications including VLMs with LMs for image captioning (\eg CLIP with GPT-2) \cite{mokady2021clipcap}, video understanding (\eg CLIP with BERT \cite{gao2021clip2tv}), visual question answering \eg \cite{song2022clip} and ALMs and LMs for speech and text modeling \eg \cite{song2022large,bapna2022mslam}. These systems are often finetuned on task-specific data, and while this paradigm is likely to be preferred in domains for which data is abundant, our results suggest that SMs can be a strong  alternative for applications in which data is less available or more expensive to obtain.

\textbf{Multimodal probabilistic inference} is an alternative \eg Bayesian approach where one model is used as a prior and the other as evidence -- with which models from different modalities may perform joint inference \cite{karpagavalli2016review,saycan2022arxiv}. One prominent example is in automatic speech recognition: different language models can be trained separately, then transfer knowledge to a speech-to-text system via priors \cite{karpagavalli2016review}.%in which $p({\color{RoyalBlue}\text{language }}| {\color{purple}\text{audio}}, {\color{RoyalBlue}\text{language}_{\text{history}}})$ may involve both a term $p({\color{RoyalBlue}\text{language}} | {\color{purple}\text{audio}})$, and a term $p({\color{RoyalBlue}\text{language}} | {\color{RoyalBlue}\text{language}_{\text{history}}})$,
%\andy{maybe directly talk about limitations here?}

% SMs present a language-based approach to combining the outputs of multiple foundation models, which differs from a classical Bayesian approach where one model is used as a prior and the other as evidence. Relying on language-only multi-model discussion carries both pros and cons. For example, the intermediate outputs of the models may be more interpretable, but are treated as ``truth'' between models -- \ie not weighing them against the other's priors or evidence, which can lead to more divergent model interactions.

%\section{Related Work}\label{sec:background-and-related}

% \textbf{Multi-model multimodal reasoning.}

The notion of ``Mixture-of-Experts'' (\cite{jordan1994hierarchical}, see \cite{masoudnia2014mixture} for a review) is also a common paradigm for combining the outputs of multiple models -- specifically, mixtures of experts across multimodal domains including vision and audio \cite{liu2019use} have been studied. % (results from Liu et al. \cite{liu2019use} included in Table~\ref{table:msr-vtt-results-mini}).
Further investigating these techniques in the context of recent pretrained foundation models may be a promising direction for future work. % Our work may be interpreted as a particular extension of Mixture-of-Experts in which experts may be composed to provide feedback to each other, closed-loop, via the common representation of language.

\textbf{Zero-shot or few-shot prompting} recently has been shown, notably by Brown et al. \cite{brown2020language}, to be highly effective for transfer learning. In this approach, a large pretrained language model is zero-shot or few-shot {\em{prompted}} with several examples, without training, to perform a new task. Further methods such as chain-of-thought prompting \cite{wei2022chain} have shown that even simple prompting modifications can have a profound impact on target task performance \cite{wei2022chain,chowdhery2022palm} and enable new capabilities. Our work builds on these works, by extending prompting methods into the multimodal domain. 

% \textbf{Foundation models? Language models? VQA?.}

% Merlot Reserve https://arxiv.org/pdf/2201.02639.pdf
% CLIP-based VQA https://arxiv.org/pdf/2203.07190.pdf

% Audio retrieval with natural language queries https://arxiv.org/pdf/2105.02192.pdf, follow up
% https://arxiv.org/pdf/2112.09418.pdf

% {\textbf{The ``Train All Multimodal Together'' Approach.}}

\section{Socratic Models}

% \begin{wrapfigure}{r}{0.5\textwidth}
%   \vspace{-1em}
%   \captionsetup{type=figure}
%   \includegraphics[width=\linewidth]{figures/idea-fig-3.png}
%   \vspace{-1.0em}
%   \captionof{figure}{\small Socratic Models are composed via guided multi-model exchanges.  In this example, {\color{RoyalBlue}LMs}, {\color{LimeGreen}VLMs}, and {\color{Purple}ALMs} are composed closed-loop via language. 
%   % Examples of guided multi-model discussions are provided in Sec.~\ref{subsec:linguistic-world-state}, Sec.~\ref{subsec:open-ended-qa}, Sec.~\ref{sec:internet-image}, and Sec.~\ref{sec:internet-video-to-text} on various multimodal applications.
%   % https://docs.google.com/drawings/d/1tVYfxG2mKC_LF3pDG63bwaJ2mV58dKGk0YFBWYzmE5w/edit?resourcekey=0-daxT05lAVGiFrx9SD9uG-w
%   }
%   \vspace{-1.0em}
%   \label{fig:idea-fig}
% \end{wrapfigure}

Socratic Models (SMs) is a framework in which multiple large pretrained models may be composed through language (via prompting) without requiring training, to perform new downstream multimodal tasks. % \eg image captioning, video-to-text retrieval, egocentric video Q\&A, multimodal assistive dialogue, and robot planning. %One common trend in multimodal learning has been to seek embedding spaces in which multiple modalities may co-exist. While we may leverage submodels built with shared embedding spaces, language is the primary representation upon which multiple of these models, as well as other language-interactable models, exchange information closed-loop to perform joint inference.
%These guided multi-model exchanges are best described through examples, as provided in Sec. \ref{sec:evaluation} and \ref{sec:applications}. % ~\ref{subsec:linguistic-world-state}, \ref{subsec:open-ended-qa}, \ref{sec:internet-image}, and ~\ref{sec:internet-video-to-text}.
%
This offers an alternative method for composing pretrained models that directly uses language as the intermediate representation by which the modules exchange information with each other. It is both distinct from, and may be complementary to, other multimodal approaches such as joint multimodal training (Sec.~2). SMs are perhaps most intuitively understood through examples, which are provided in Sec.~\ref{sec:evaluation} and \ref{sec:applications}, but a definition is as follows.
A task-specific Socratic Model $f_{\text{SM}}: \mathcal{X} \rightarrow \mathcal{Y}$ may be described as a computation graph, with nodes as a set of modules $\{f^i_{\mathcal{M}^i}\}$, and the edges of the graph represent intermodule communication through language.
% $f^1_{\mathcal{M}^1}$ to $f^N_{\mathcal{M}^N}$:
%$$f_{\text{SM}} = f^N_{\mathcal{M}^N}( \ldots f^2_{\mathcal{M}^2}( f^1_{\mathcal{M}^1}(\mathcal{X})))$$
Each $\mathcal{M}$ is some (multimodal) model or external API, and each module $f$ assists in transforming the output of one $f$ into a form of language that a connected $f'$ may use for further inference. For visualization, outputs from LMs are {\color{RoyalBlue}blue}, VLMs {\color{LimeGreen}green}, ALMs {\color{Purple}purple}, prompt text {\color{Gray}gray}, user inputs {\color{magenta}magenta}, VLM-chosen LM outputs {\greenuline{green-underlined blue}}, and ALM-chosen LM outputs {\purpleuline{purple-underlined blue}}.

% A task-specific Socratic Model $f_{\text{SM}}: \mathcal{X} \rightarrow \mathcal{Y}$ may be described as the sequential composition of modules $f^1_{\mathcal{M}^1}$ to $f^N_{\mathcal{M}^N}$:
% $$f_{\text{SM}} = f^N_{\mathcal{M}^N}( \ldots f^2_{\mathcal{M}^2}( f^1_{\mathcal{M}^1}(\mathcal{X})))$$
% where each $\mathcal{M}$ is some (multimodal) model or external API, and each module $f$ assists in transforming the output of one $f^i$ into a form of language that $f^{i+1}$ may use for further inference. For visualization, outputs from LMs are {\color{RoyalBlue}blue}, VLMs {\color{LimeGreen}green}, ALMs {\color{Purple}purple}, prompt text {\color{Gray}gray}, user inputs {\color{magenta}magenta}, VLM-chosen LM outputs {\greenuline{green-underlined blue}}, and ALM-chosen LM outputs {\purpleuline{purple-underlined blue}}.

\begin{wrapfigure}{r}{0.27\textwidth}
  \vspace{-1.1em}
  \begin{subfigure}{0.27\textwidth}
  \centering
  \includegraphics[width=\linewidth]{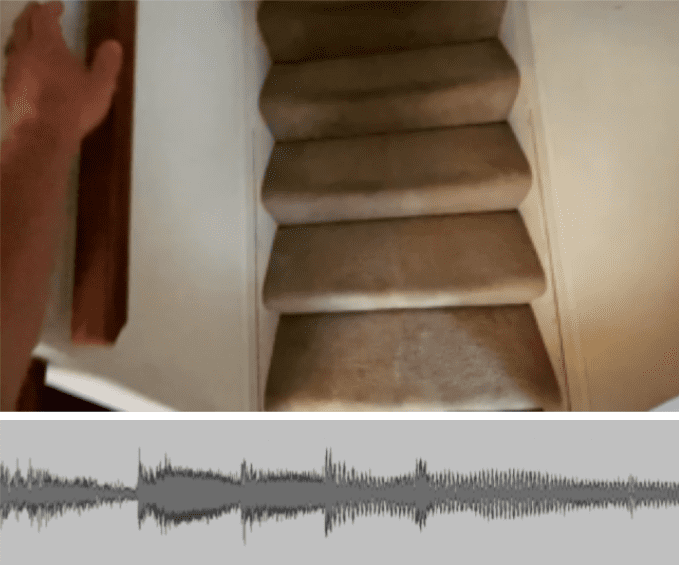}
  \end{subfigure}%
  
  \begin{subfigure}{0.27\textwidth}
  \centering
  \noindent\fcolorbox{lightgray}{white}{%
    \parbox{0.93\linewidth}{%
      \vspace{-0.5em}
      \begin{flushleft}\scriptsize{
        \texttt{{\color{Gray}I am in a: {\color{LimeGreen}staircase}. I see a: {\color{LimeGreen}stairs, animal, mammal, hamster, human leg}. I think I hear {\purpleuline{footsteps}}. I am: {\greenuline{climbing}}. Summary: {\color{RoyalBlue}I am most likely climbing a staircase, and I may hear footsteps}.}}}
        \vspace{-0.5em}
      \end{flushleft}
    }%
  }
  \end{subfigure}%
  \vspace{-0.2em}
  
  % \captionsetup{type=figure}
  \caption{\small Example: SM systems can be prompted to zero-shot annotate an egocentric image with a summary of the person's activities. Information from multiple modalities ({\color{RoyalBlue}language}, {\color{Purple}audio}) can help denoise predictions from any one specific modality ({\color{LimeGreen}vision}).}
  \label{fig:ego-audio}
  \vspace{-1.5em}
\end{wrapfigure}

A key component in SMs is {\em{multi-model multimodal prompting}}, in which information from a non-language domain is substituted into a language prompt, which can be used by an LM for reasoning. One way to multimodal prompt is to variable-substitute language-described entities from other modalities into a prompt.  An example of this is: $\text{activity} = f_\text{\color{RoyalBlue}LM}(f_\text{\color{LimeGreen}VLM}(f_\text{\color{RoyalBlue}LM}(f_\text{\color{Purple}ALM}(f_\text{\color{RoyalBlue}LM}(f_\text{\color{LimeGreen}VLM}(\text{video}))))))$ shown in Fig.~\ref{fig:ego-audio}, where (i) the VLM detects visual entities, (ii) the LM suggests sounds that may be heard, (iii) the ALM chooses the most likely sound, (iv) the LM suggests possible activities, (v) the VLM ranks the most likely activity, (vi) the LM generates a summary of the Socratic interaction.
Some form of such multimodal prompting is central to all of our demonstrated SM examples (Sec.~\ref{sec:evaluation} and \ref{sec:applications}). Note that this example involves multiple back-and-forth interactions, including calling the same model multiple times, forming a sort of ``closed-loop'' feedback between nodes in the SM graph.

Informally SMs may be interpreted as composing pretrained models to ``talk to each other'', but in practice certain models may need simple pre- and post-processing to produce language. For example, vision-text similarity VLMs, \eg CLIP \cite{radford2021learning}, do not inherently produce text, but can be made to perform zero-shot detection from a large pre-existing library of class category names, and return the top-$k$ detected categories. Accordingly, although our example SM systems required no training, the interactions between models are scripted with prompt templates.
While in future work we are excited to explore learning the interactions (\ie forms of each $f$, and edges), we also find
practical benefits of a framework with no required task-specific training: new applications can be quickly targeted by just a small amount of creative programming.

SMs are in part a reaction to the constraints of the predominant ``pretraining weights'' (Sec.~\ref{sec:background-and-related}) paradigm to transfer learning with foundation models, which include: (i) expensive (at times prohibitively) to finetune large 100B+ parameter models both in terms of compute costs and data collection (can be challenging for new multimodal applications \eg in AR or robotics), (ii) finetuning pretrained model weights may lose generality and robustness to distribution shifts \cite{wortsman2021robust}, (iii) foundation models may store ``stale'' knowledge (due to training latencies), and lack access to dynamic online data or proprietary sources of information. Despite these limitations, large pretrained foundation models \cite{bommasani2021opportunities} are likely to serve as a backbone for many intelligent systems of the future -- SMs is a systems approach (\ie glue framework) that leans on their zero-shot and few-shot capabilities in aggregate as a means to address these limitations for new downstream multimodal tasks.

\section{Evaluation: Methods and Results}\label{sec:evaluation}

In this section, we quantitatively evaluate Socratic Models on: image captioning \cite{chen2015microsoft, lin2014microsoft} (Sec. 3.1), contextual image captioning \cite{kreiss2021concadia} (Sec. 3.2), and video-to-text retrieval \cite{xu2016msr} (Sec. 3.3). For each task, we (i) describe how we use the SMs framework, and (ii) discuss results.

\subsection{Socratic Image Captioning on MS COCO Captions: {\color{LimeGreen}VLM} + {\color{RoyalBlue}LM} }\label{sec:mscoco-captioning}
  %We describe an example system in Sec.~\ref{subsec:internet-captioning-overview} and demonstrate results in Sec.~\ref{subsec:internet-captioning-results}.

%\subsection{System: Image Captioning on Internet Data}\label{subsec:internet-captioning-overview}

\begin{figure}[H]
\vspace{-0.5em}
\centering
\begin{subfigure}{.27\textwidth}
  \vspace{-0.9em}
  \centering
  \noindent\fbox{%
    \parbox{0.91\linewidth}{%
      \vspace{-0.5em}
      \begin{flushleft}\scriptsize{
        \texttt{{\color{Gray}I am an intelligent image captioning bot.
        This image is a {\color{LimeGreen}\{img\_type\}}. There {\color{LimeGreen}\{num\_people\}}.
        I think this photo was taken at a {\color{LimeGreen}\{place1\}}, {\color{LimeGreen}\{place2\}}, or {\color{LimeGreen}\{place3\}}.  I think there might be a {\color{LimeGreen}\{object1\}}, {\color{LimeGreen}\{object2\}}, {\color{LimeGreen}\{object3\}},... in this {\color{LimeGreen}\{img\_type\}}.
        A creative short caption I can generate to describe this image is:}}}
        \vspace{-0.5em}
      \end{flushleft}
    }%
  }
  % \caption{VLM + LM prompt.}
  % \label{fig:sub1}
\end{subfigure}%
\begin{subfigure}{.73\textwidth}
  \centering
  \includegraphics[width=1.0\linewidth]{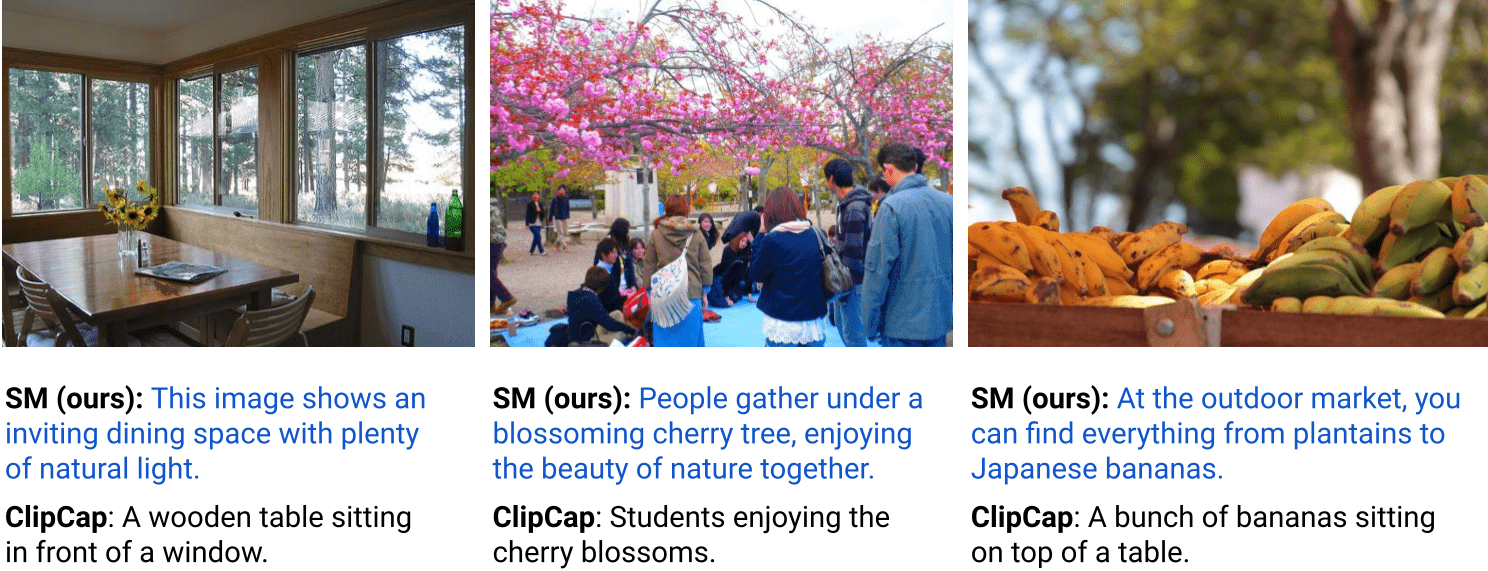}
  % \caption{Example results on MS COCO images.}
  % \label{fig:sub2}
\end{subfigure}
% \vspace{-0.5em}
\caption{\small SMs with {\color{LimeGreen}VLM} and {\color{RoyalBlue}LM} prompting (left) can zero-shot generate captions for generic Internet images (\eg from MS COCO), and can be as expressive as task-specific finetuned methods such as ClipCap \cite{mokady2021clipcap}.}
\label{fig:SM-caption-small}
\vspace{-1.5em}
\end{figure}

\textbf{Method.}
SMs can generate image captions by prompting a guided language-based exchange between a VLM and LM -- \ie via $\text{caption} = f^3_{\text{\color{LimeGreen}VLM}}(f^2_{\text{\color{RoyalBlue}LM}}(f^1_{\text{\color{LimeGreen}VLM}}(\text{image})))$.
% Generating image captions can be formulated with SMs as a guided multi-model exchange between a VLM and LM.
%Overall, our example SMs system for Internet image captioning is extremely similar to how we perform single-image captioning in our egocentric system, but (i) adapted for Internet images rather than ego-centric images, and (ii) adapted such that the ``final task'' is the generation of a single image caption, rather than open-ended tasks based on text-prompted completion.
% First, we prompt the VLM to zero-shot detect visual entities across different categories of language -- our example system detects across visual entities across
First (1), the {\color{LimeGreen}VLM} is used to zero-shot detect different place categories (Places356 \cite{zhou2016places}), object categories (from Tencent ML-Images \cite{wu2019tencent}), image type ({\{\em{photo, cartoon, sketch, painting}\}}) and the number of people \{{\em{are no people, is one person, are two people, are three people, are several people, are many people}}\}.
% by substituting the category name into the phrase ``Photo of X.'' and rank matching it against images
%For generic Internet images, which are not necessarily real photos, and very often are taken by people of people, we find that these additional contextual pieces of information help generate better captions.
%In this example, VLM detections give: ``Places: {\color{LimeGreen}\{place1\}}, {\color{LimeGreen}\{place2\}}, {\color{LimeGreen}\{place3\}}. Objects: {\color{LimeGreen}\{object1\}}, {\color{LimeGreen}\{object2\}}, {\color{LimeGreen}\{object3\}}. Image type: {\color{LimeGreen}\{image\_type\}}. People result: {\color{LimeGreen}\{people\_result\}}.''
The top-$k$ ranked in each category can then be substituted into an LM prompt, as shown in Fig.~\ref{fig:SM-caption-small}, left. Second (2), given the VLM-informed language prompt, a causal {\color{RoyalBlue}LM} (\ie for text completion) generates several $n$ candidate captions. For this step, we use a non-zero next-token
sampling temperature (\eg 0.9 for GPT-3), to return sufficiently diverse, but reasonable results across the $n$ candidates.  %Unlike in the ego-centric perception system, our goal is generate specifically generate single-image captions.
Finally (3), these $n$ captions are then ranked by the {\color{LimeGreen}VLM} with the image, and the highest scoring caption is returned.

\begin{wraptable}{r}{0.49\textwidth}
% \begin{table*}[h]
  \vspace{-1.0em}
  \setlength\tabcolsep{2.0pt}
  \scriptsize
  \begin{tabular}{@{}lcccccc@{}}
  \toprule
  Method         &  BLEU-4 & METEOR &  CIDEr & SPICE & ROUGE-L \\
  \midrule
  $^*$ClipCap \cite{mokady2021clipcap} & 40.7 & 30.4 & 152.4 & 25.2 & 60.9\\
  % \multirow{2}{*}{{\em{Finetuned (unpaired)}}} & \multirow{2}{*}{ } & \multirow{2}{*}{ } & \multirow{2}{*}{ } & \multirow{2}{*}{ } & \multirow{2}{*}{ } & \multirow{2}{*}{ } &
  $^\dagger$MAGIC \cite{su2022language} & 11.4 & 16.4 & 56.2 & 11.3 & 39.0\\
  \midrule
  ZeroCap \cite{tewel2021zero} & 0.0 & 8.8 & 18.0 & 5.6 & 18.3\\
  SMs 0-shot (ours) & 6.9 & 15.0 & 44.5 & 10.1 & 34.1\\
  SMs 3-shot (ours) & \textbf{18.3} & \textbf{18.8} & \textbf{76.3} & \textbf{14.8} & \textbf{43.7}\\
  \bottomrule
  \end{tabular}
  
  \vspace{0.3em}
  $^*$finetuned on full training set with image-text pairs.\\
  $^\dagger$finetuned on unpaired training set, zero-shot on image-text pairs.
  \vspace{-0.5em}
  \caption{\small{Image captioning comparisons on a random subset of $N=100$ MS COCO test examples.
  }}
  \vspace{-2.0em}
  \label{table:mscoco-results}
  % table: https://docs.google.com/document/d/1nPJH1DkaBN7vqX12b3bxAszbW-YyJm4AEXrDmaZuig0/edit#
% \end{table*}
\end{wraptable}

\textbf{Results.} Tab.~\ref{table:mscoco-results} shows quantitative comparisons with state-of-the-art image captioning methods on MS COCO Captions dataset \cite{chen2015microsoft,lin2014microsoft}. We chose to evaluate over a random sampled subset of 100 images from the test split \cite{karpathy2015deep}, so that GPT-3 API runtime costs are more affordable for reproducibility ($\sim$\$150 USD per run with with $n=20$ generated candidate captions per image). Metrics from baselines are comparable to full-test-set metrics (see Appendix).

SMs substantially outperform the zero-shot state-of-the-art ZeroCap \cite{tewel2021zero} with a CIDEr \cite{vedantam2015cider} score 18.0 $\rightarrow$ 44.5, but do not perform as well as methods such as ClipCap \cite{mokady2021clipcap} which are directly finetuned on the training set. SMs tend to generate verbose and descriptive captions (see qualitative examples in Fig.~\ref{fig:SM-caption-small}), but may naturally score lower on captioning metrics if they do not match the dataset's distribution of caption labels. This performance gap narrows as SMs are few-shot prompted with 3 random captions from the training set, bringing CIDEr scores up to 76.3, exceeding the performance of MAGIC \cite{su2022language} which finetunes the text generator on the training set's unpaired captions.
% For reference, Human level performance is about 0.9 \cite{chen2015microsoft,lin2014microsoft}. since there are many different ways to describe an image.

% \textbf{(\textit{3.1.ii}) Results.} Fig.~\ref{fig:clipcap-comparisons} shows several qualitative comparisons with ClipCap \cite{mokady2021clipcap}, a state-of-the-art method for image captioning specifically trained via finetuning on image captioning. We show zero-shot generation of captions on the set of images displayed in ClipCap's code release.\footnote{\href{https://github.com/rmokady/CLIP\_prefix\_caption}{https://github.com/rmokady/CLIP\_prefix\_caption}} We make a colab available to generate these results.\footnote{Note that due to the non-zero temperature used for sampling from the generative language model, results from this approach are stochastic, but comparable results are producible.}

% Overall, our results (Fig.~\ref{fig:clipcap-comparisons}) show that the Socratic Model framework can be adopted to provide often convincing results for image captioning via the creativity of the LM combined with the visual grounding of the VLM.
While these results are promising, the degree to which visual details are provided in the captions is largely limited by the capabilities of the VLM. For example, attributes (\eg color of a shirt, a person's facial expression, or the spatial relationships between objects) are details not often captured in our particular system, which relies more on the contextual image classification capabilities of the VLM. Future captioning work with SMs may explore open-vocabulary object detectors \cite{gu2021open,kamath2021mdetr} to recover more salient details, or combine the outputs of multiple task-specific image captioning models with LMs to assemble a single rich and coherent caption.

\subsection{Socratic Contextual Image Description on Concadia:  {\color{LimeGreen}VLM} + {\color{RoyalBlue}LM} }\label{sec:concadia-captioning}

\textbf{Method.} Concadia \cite{kreiss2021concadia} is a dataset for contextual image captioning and description, conditioned on the input image and an associated paragraph of article text. In particular, image descriptions describe the visual content in the image (\eg ``a portrait of a man with a beard in a suit'') commonly used for accessibility, while image captions link images to article text (\eg ``a photo of Abraham Lincoln''). We evaluate SMs on both tasks, using a similar method to MS COCO captions (Sec. \ref{sec:mscoco-captioning}) but with article-text prompt-substitution (below), and no need for VLM re-ranking. $f^2_{\text{\color{RoyalBlue}LM}}(f^1_{\text{\color{LimeGreen}VLM}}(\text{image}),\text{context})$:

\noindent\fbox{%
    \parbox{0.97\linewidth}{%
        \scriptsize{\texttt{{\color{Gray}I am an intelligent image captioning bot. The article is about: "{\color{Black}\{article\_text\}}". In this image, I think I see a {\color{LimeGreen}\{object1\}}, {\color{LimeGreen}\{object2\}}, {\color{LimeGreen}\{object3\}},... A short caption for this image is:}}}
    }%
}\\

\begin{wraptable}{r}{0.47\textwidth}
  \vspace{-1.0em}
  \setlength\tabcolsep{2.0pt}
  \scriptsize
  \begin{tabular}{@{}lccc@{}}
  \toprule
  Method         &  Caption Generation & Description Generation \\
  \midrule
  Kreiss et al. \cite{kreiss2021concadia} & 11.3 & 17.4\\
  % SMs (no image) & \textbf{40.1} & 20.6\\
  SMs (ours) & \textbf{38.9} & \textbf{22.6}\\
  {\color{Gray}SMs w/ description} & {\color{Gray}93.8} & {\color{Gray}--}\\
  \bottomrule
  \end{tabular}
  \caption{\small{SMs zero-shot are competitive on contextual image captioning and description (measured with CIDEr) on the Concadia dataset, outperforming task-specific methods \eg Kreiss et al. \cite{kreiss2021concadia} which finetunes on the training set.
  }}
  \vspace{-1.0em}
  \label{table:concadia-results}
\end{wraptable}

\textbf{Results.}
We evaluate SMs on the test split of Concadia with 9,691 images (shown in Tab. \ref{table:concadia-results}).
Despite being zero-shot, SMs outperform the task-specific prior best method, Kreiss et al. \cite{kreiss2021concadia}, that directly finetunes on the training set of 77,534 images, with a CIDEr score improvement 11.3~$\rightarrow$~38.9 for generated image captions, and 17.4~$\rightarrow$~22.6 for generated image descriptions.
We also report numbers for captioning generation conditioned on the image, article text, and ground truth description. This achieves a CIDEr score of 93.8 and suggests the upper bound of performance if SMs are used with VLMs that can produce accurate image descriptions.
We also discuss interesting additional findings in the appendix, \eg that LMs can perform comparably on contextual image captioning even without input images (\ie only article text as input), which either (i) reflects a strong correlation between the distributions of captions and article texts, and/or (ii) indicates LM training set overlap. Overall, the results on Concadia are promising and suggest that SMs can be used to automatically generate descriptive texts that improve the accessibility of visual content on the web for the low vision community. 

\subsection{Socratic Video-to-Text Retrieval: {\color{LimeGreen}VLM} + {\color{RoyalBlue}LM} + {\color{Purple}ALM}}\label{sec:internet-video-to-text}

% We also adapt the Socratic Models framework to the task of video-to-text retrieval, a common video understanding task with clear and quantitative metrics.  We describe an example SM-based system (Sec.~\ref{subsec:video-to-text-system}) which uses a guided multi-model exchange between a VLM, a causal LM, a speech-to-text (ALM) model, and a masked LM, that achieves state-of-the-art performance for zero-shot video-to-text (\ie caption) retrieval on the popular MSR-VTT dataset (Sec.~\ref{subsec:video-to-text-results}).

\textbf{Method.} Socratic Models can be adapted for video-to-text retrieval, a video understanding task commonly benchmarked on MSR-VTT \cite{xu2016msr}.
Our approach leverages commonsense information from audio and language domains to augment the vision-based Portillo-Quintero et al. \cite{portillo2021straightforward}, which computes a similarity measure between the average VLM (\ie CLIP) features of all video frames per video, and the VLM text features of captions -- used to execute video-to-text retrieval with one-to-many nearest neighbor matching.
Specifically, our system transcribes audio from the video with speech-to-text ALMs \cite{bapna2022mslam} for automatic speech recognition (ASR \eg via Google Cloud speech-to-text API \cite{gcloud-speech-to-text}), then summarizes the transcripts with an LM using the following prompt:

\noindent\fbox{%
    \parbox{0.97\linewidth}{%
        \scriptsize\texttt{{\color{Gray}
I am an intelligent video captioning bot.'
I hear a person saying: "{\color{Purple}\{transcript\}}".
Q: What's a short video caption for this video? A: In this video,
        }
    }%
}}

% Portillo-Quintero et al. \cite{portillo2021straightforward} computes a similarity measure between the average VLM (\ie CLIP) features of the image frames from a video, and the VLM text features of a caption. 
% This zero-shot method can be used directly for video-to-text retrieval via one-to-many nearest neighbor matching.  
% We can improve upon their method through a combination of speech-to-text ALMs together with LM-based commonsense reasoning.
%Although raw transcripts may be challenging to incorporate into meaningful improvements for video/caption retrieval, we may leverage reasoning capabilities from large LMs in order to usefully harness the transcripts.
%For videos with sufficiently long transcripts, we summarize the content with an LM (\eg GPT-3) using the following prompt:

We compute similarity scores of the generated summary to the set of captions with a masked LM (\eg similarity between sentence embeddings from RoBERTa \cite{liu2019roberta}), and use those scores to re-weight the CLIP-based ranking from Portillo-Quintero et al. For videos with sufficiently-long transcripts ($\ge$100 characters), the matching score is: $\big( {\color{LimeGreen}\textit{CLIP~}}(\text{caption}) \boldsymbol{\cdot} {\color{LimeGreen}\textit{CLIP~}}(\text{video}') \big) \times \big( {\color{RoyalBlue}\textit{RoBERTa~}}(\text{caption}) \boldsymbol{\cdot}  {\color{RoyalBlue}\textit{RoBERTa~}} ({\color{RoyalBlue}\textit{GPT-3}}({\color{Gray}prompt},~{\color{Purple}\textit{Speech2Text~}}(\text{audio}')))\big)$, where $\boldsymbol{\cdot}$ represents normalized dot product of embeddings, and $\times$ represents scalar multiplication. For a given video, if there is no audio or the transcript is too short, we default to Portillo-Quintero et al., which is just ${\color{LimeGreen}\textit{CLIP}}(\text{caption}) \boldsymbol{\cdot} {\color{LimeGreen}\textit{CLIP}}(\text{video}')$. Here, the Socratic interaction lies mainly between the ALM (speech-to-text) to the commonsense LM (GPT-3 to summarize transcriptions), and between the commonsense LM to the ranking based system that is a combination of the VLM (CLIP) and the masked LM (RoBERTa). 

\begin{wraptable}{r}{0.5\textwidth}
% \begin{table*}[h]
  \vspace{-1.0em}
  \setlength\tabcolsep{1.0pt}
  \centering
  \scriptsize
  \begin{tabular}{@{}llccccccccll@{}}
  \toprule
  \multicolumn{2}{l}{} & \multicolumn{4}{c}{MSR-VTT Full} &  &  \\
  \cmidrule(lr){3-6}
  Category & Method         &  R@1$\uparrow$ & R@5$\uparrow$ &  R@10$\uparrow$ & MdR$\downarrow$    & Audio \\
  \midrule
  \multirow{3}{*}{\em{Finetuned}} & JEMC \cite{mithun2018learning}  & 12.5 & 32.1  & 42.4 & 16.0 & yes &  \\
                                   & Collab. Experts \cite{liu2019use}  & 15.6 & 40.9 & 55.2 & 8.3 & yes &  \\
                                   & CLIP2Video \cite{fang2021clip2video}  & \textbf{54.6} & \textbf{82.1} & \textbf{90.8} & \textbf{1.0} & no \\
  \midrule
  \multirow{2}{*}{\em{Zero-shot}}  & CLIP via \cite{portillo2021straightforward} & 40.3 & 69.7 & 79.2 & \textbf{2.0} & no\\
                                   & SMs (ours)  & \textbf{44.7}  &  \textbf{71.2} &  \textbf{80.0} & \textbf{2.0} & yes\\
  \bottomrule
  \end{tabular}
  % \vspace{-0.5em}
  \caption{\small{Video-to-text retrieval results on MSR-VTT \cite{xu2016msr} dataset, both on the popular 1k-A \cite{yu2018joint} subset and the original `full' test set.  Differentiated are methods which train on the MSR-VTT dataset ({\em{finetuning}}), compared with {\em{zero-shot}} methods, which do not.  Also noted: whether the methods use audio channels, and if CLIP \cite{radford2021learning} is used, which CLIP encoder is used. % See Fig.~\ref{fig:msr-vtt-sota} for the chronology of the SOTA across each category.
  }}
  \vspace{-1.0em}
  \label{table:msr-vtt-results-mini}
  % table: https://docs.google.com/document/d/1nPJH1DkaBN7vqX12b3bxAszbW-YyJm4AEXrDmaZuig0/edit#
% \end{table*}
\end{wraptable}

\textbf{Results.} We evaluate on MSR-VTT \cite{xu2016msr}, which as noted in other recent works \cite{gao2021clip2tv,cheng2021improving} is the most popular benchmark for video-to-text retrieval.
We compare our method with both zero-shot methods, as well as finetuned methods specifically trained on MSR-VTT. Results show that our method sets a new zero-shot state-of-the-art (Tab.\ref{table:msr-vtt-results-mini}). Since our system uses Portillo-Quintero et al. \cite{portillo2021straightforward} to process CLIP features but additionally incorporates LM reasoning on speech-to-text transcripts, the increased measured performance of our method (\ie 40.3 $\rightarrow$ 44.7 R@1) directly reflects the added benefits of incorporating language-based multimodal reasoning.  Additionally, to keep the comparison between our method and Portillo-Quintero et al. \cite{portillo2021straightforward} as direct as possible, we maintain the usage of their precomputed CLIP features from ViT-B/32, but it is likely that performance can be improved with other recent more performant VLMs (\eg LiT \cite{zhai2021lit}, CLIP with ViT-L/14).
%Given results from other recent methods (Tab.~\ref{table:msr-vtt-results-mini}), it is likely that performance can be improved by switching to ViT-B/16, or other recent more performant VLM models \cite{zhai2021lit}.

\begin{wraptable}{r}{.35\textwidth}
% \begin{table}[h]
  \vspace{-1.0em}
  \setlength\tabcolsep{2.0pt}
  \centering
  \scriptsize
  \begin{tabular}{@{}lcccc@{}}
  \toprule
  & \multicolumn{4}{c}{\em{Long-transcript subset of}}  \\
  & \multicolumn{4}{c}{MSR-VTT Full}   \\
  \cmidrule(lr){2-5}
               & R@1$\uparrow$ & R@5$\uparrow$ & R@10$\uparrow$ & MdR$\downarrow$ \\
  \midrule
  CLIP via \cite{portillo2021straightforward}          & 41.5 & 69.6 & 77.4 & 2.0  \\
  SMs (ours)           & \textbf{54.9} & \textbf{74.0} & \textbf{79.9} & \textbf{1.0}  \\
  \bottomrule
  \end{tabular}
  %\vspace{-0.5em}
  \caption{\small{SMs substantially improve on Portillo-Quintero et al. \cite{portillo2021straightforward} for video-to-text retrieval on the MSR-VTT subset of videos for which {\em{long-transcripts}} are available ($n$=1,007 out of 2,990).
  % On these subsets, we evaluate  vs. our method.  Outside this subset, we resort to Portillo-Quintero et al.
  }} % task, success \% vs. $\#$ of demonstrations. 
  \vspace{-1.0em}
  \label{table:msr-vtt-subset}
  % table: https://docs.google.com/document/d/1nPJH1DkaBN7vqX12b3bxAszbW-YyJm4AEXrDmaZuig0/edit#
% \end{table}
\end{wraptable}

Table~\ref{table:msr-vtt-subset} shows that on the subset of test videos that contain {\em{long-transcripts}}, we observe a more substantial increase in performance from 40.3 to 54.9 with our method compared to Portillo-Quintero et al. \cite{portillo2021straightforward}. Note that this is roughly comparable to the R@1 of the best {\em{finetuned}}-SOTA method, CLIP2Video \cite{fang2021clip2video}, with 54.6 R@1 (Tab.~\ref{table:msr-vtt-results-mini}).
If we assume that for visual-only methods, the videos with-or-without transcripts are of roughly equal difficulty from a visual-only retrieval perspective, this suggests that on internet videos with sufficient speech present in the audio, our zero-shot SMs can nearly match the {\em{finetuned}}-SOTA methods for video-to-text retrieval.

\section{Applications: Methods and Demonstrations}\label{sec:applications}

In this section, we describe several applications of SMs on (i) egocentric perception, (ii) multimodal assistive dialogue, and (iii) robot perception and planning. These applications each involve processing user inputs/feedback, and serve as examples of integrating external modules (\eg web search, robot policies) as additional participants to a Socratic discussion to enable new multimodal functionalities.

\subsection{Egocentric Perception: {\color{magenta}User} + {\color{LimeGreen}VLM} + {\color{RoyalBlue}LM} + {\color{Purple}ALM}}\label{subsec:system-overview}

SMs can be prompted to perform various perceptual tasks on egocentric video: (i) summarizing content, (ii) answering free-form reasoning questions, (iii) and forecasting. Egocentric perception has downstream applications in AR and robotics, but remains challenging: the characteristics of first-person footage -- from unusual viewpoints to lack of temporal curation -- are not often found in existing datasets, which focus more on generic Internet content captured from third-person views \cite{deng2009imagenet,lin2014microsoft,sharma2018conceptual}. This domain shift makes it difficult for data-driven egocentric models to benefit from the paradigm of pretraining on third person Internet data \cite{li2021ego,sigurdsson2018charades}. SMs offer a zero-shot alternative to perform egocentric perceptual tasks without training on large domain-specific datasets \cite{grauman2021ego4d,damen2020rescaling,sigurdsson2018charades}.

For open-ended reasoning, a key aspect of our SMs-based approach is formulating {\em{video understanding as reading comprehension}}, \ie re-framing ``video Q\&A'' as a ``short story Q\&A'' problem, which differs from common paradigms for video understanding that may involve supervising video-text models on labeled datasets or adversarial training (see \cite{patel2021recent} for a recent survey). To this end, we first extract a set of ``key moments'' throughout the video (\eg via importance sampling, or video/audio search based on the input query, discussed in Appendix). We then caption the key frames indexed by these moments (using prompts similar to those in Sec. \ref{sec:mscoco-captioning} and Sec. \ref{sec:concadia-captioning}), and recursively summarize \cite{wu2021recursively} them into a language-based record of events, which we term a {\em{language-based world-state history}}. This is then passed as context to an LM to perform various reasoning tasks via text completion such as Q\&A, for which LMs have demonstrated strong zero-shot performance \cite{brown2020language}. Drawing analogies to 3D vision and robotics, the world-state history can be thought of as building an on-the-fly reconstruction of events in the observable world with language, rather than other representations, such as dynamically-updated 3D meshes \cite{izadi2011kinectfusion} or neural fields \cite{tancik2022block}.

% wav2clip?

\textbf{(i) Summarization} enables augmenting human memory to recall events or life-log activities. Given world-state history constructed from SMs using a first-person POV video\footnote{\small Examples on \href{https://youtu.be/-UXKmqBPk1w}{https://youtu.be/-UXKmqBPk1w} used with permission from Cody Wanner.}, this can be implemented by prompting an LM to complete: ``{\{\greenuline{world-state history}\}} {\color{Gray} Summary of my day:}'' to which it can respond with outputs like ``{\color{RoyalBlue}I slept in a bed, made coffee, watched TV, did laundry, received a package, bench pressed, showered, ate a sandwich, worked on a computer, and drank wine.}''

\textbf{(ii) Open-ended Q\&A} involves prompting the LM to complete the template: ``{\{\greenuline{world-state history}\}} {\color{Gray}Q:} {\color{magenta}\{question\}} {\color{Gray}A:}''. Conditioned on the quality (comprehensiveness) of the world-state history, LMs can generate surprisingly meaningful results to contextual recall questions (\eg ``{\color{magenta}what was I doing outdoors?}''~$\rightarrow$~``{\color{RoyalBlue}I was chopping wood in a yard.}'', ``{\color{magenta}did I drive today?}''~$\rightarrow$~``{\color{RoyalBlue}no, I did not drive today.}''), temporal questions (\eg ``{\color{magenta}when did I last drink coffee?}''~$\rightarrow$~``{\color{RoyalBlue}I last drank coffee at 10:17 AM}'', ``{\color{magenta}how many times did I receive a package today?}''~$\rightarrow$~``{\color{RoyalBlue}I received a package once today.}''), cause-and-effect questions (\eg ``{\color{magenta}why did I go to the front porch today?}''~$\rightarrow$~``{\color{RoyalBlue}I went to the front porch today to receive a package.}'').
As in \cite{yang2021empirical} we can also further prompt the LM to \textit{explain the answer} by adding ``{\color{Gray}This is because:}'' to which it can respond ``{\color{RoyalBlue}I saw on the porch a package and knew that I was expecting it.}''

\textbf{(iii) Forecasting} of future events can be formulated as language-based world-state {\em{completion}}. Our system prompts the LM to complete the rest of an input event log. Timestamps of the predictions can be preemptively specified depending on the application needs. The completion results (example below on the right) are generative, and are more broad than binary event classification \cite{lei2020more}.

\begin{wrapfigure}{r}{0.41\textwidth}
    % \vspace{-1.1em}
    \noindent\fbox{%
        \parbox{0.97\linewidth}{%
            \scriptsize\texttt{{\color{Gray}1:46 PM: I am eating a sandwich in a kitchen.\\
    2:18 PM: I am checking time and working on a laptop in a clean room. 2:49 PM: I am buying produce from a grocery store or market.\\
    3:21 PM:} {{\color{RoyalBlue} I am driving a car.\\
    4:03 PM: I am in a park and see a playground.\\
    4:35 PM: I am in a home and see a television.}}}
        }%
    }
    \vspace{-1.0em}
\end{wrapfigure}
Few-shot prompting the LM with additional examples of prior event logs most similar to the current one is likely to improve the precision of the results, which may be useful for assistive AR applications. Without additional context, the completions are likely biased towards typical schedules seen by the LM across Internet-scale data.

% \footnote{\small Examples on https://youtu.be/-UXKmqBPk1w used with permission from Cody Wanner.}

\subsection{Multimodal Assistive Dialogue: {\color{magenta}User} + {\color{LimeGreen}VLM} + {\color{RoyalBlue}LM} + {\color{Rose}Web Search}}

SMs can be adapted to engage in multimodal dialogue to assist people in doing every day tasks, such as cooking. Our example application here helps the user search for a recipe, then guides them through it step by step. The system allows the user to navigate recipe steps with casual dialogue, provides ingredient replacements or advice (using LM priors), and searches for visual references (in the form of images or videos) on user request. This is a case study in (i) prompting a dialogue LM \cite{thoppilan2022lamda} to produce key phrase tokens that elicit specific Socratic interactions (\eg video search via a VLM to output visual data), and (ii) using a web crawler (outputs in {\color{Rose}red}) as an additional module engaged in Socratic discussion with other models to retrieve information online. The approach preconditions an LM (\eg GPT-3 \cite{brown2020language}) with context that includes when and how key phrases should be referenced:

\noindent\fbox{%
    \parbox{0.97\linewidth}{%
        \scriptsize\texttt{{\color{Gray}
Alice is a an expert chef that will help Bob prepare a given recipe. If Bob asks for the next step, Alice will respond with "Step: " followed by the next step of the recipe. If Bob does not have the right ingredients, Alice will assist Bob in finding suitable replacements. If Bob asks Alice to describe something that is better shown visually, Alice will say "(image)" followed by a response. Ingredients: {\color{Rose}\{ingredients\}} Recipe: {\color{Rose}\{recipe\}}
        }
    }%
}}

where phrases such as ``{\color{RoyalBlue}(image)}'' initiate a Socratic interaction where the web crawler returns a video of the recipe (if it exists), and the VLM runs visual search on it using the LM response to retrieve a related frame. We find that this can provide qualitatively meaningful guidance to a user, for example\footnote{Example using recipe steps and ingredients from \href{https://tasty.co/recipe/strawberry-cheesecake-macarons}{tasty.co/recipe/strawberry-cheesecake-macarons}}:
\vspace{-2.0em}
\begin{figure}[H]
\centering
\begin{subfigure}{.775\textwidth}
  \centering
%   \fbox{%
%     \parbox{0.91\linewidth}{%
%       \begin{flushleft}\scriptsize{
%         \texttt{{\color{Gray}I am an intelligent image captioning bot.
%         This image is a {\color{LimeGreen}\{img\_type\}}. There {\color{LimeGreen}\{num\_people\}}.
%         I think this photo was taken at a {\color{LimeGreen}\{place1\}}, {\color{LimeGreen}\{place2\}}, or {\color{LimeGreen}\{place3\}}.  I think there might be a {\color{LimeGreen}\{object1\}}, {\color{LimeGreen}\{object2\}}, {\color{LimeGreen}\{object3\}},... in this {\color{LimeGreen}\{img\_type\}}.
%         A creative short caption I can generate to describe this image is:}}}
%       \end{flushleft}
%     }%
%   }
  \noindent\fbox{%
    \parbox{0.97\linewidth}{%
      \vspace{-0.5em}
      \begin{flushleft}\scriptsize{
        \texttt{{\color{Gray}
Alice: {\color{RoyalBlue}Step: In a medium bowl, beat the egg whites until frothy.}\\
Bob: {\color{magenta}how many eggs.} \\
Alice: {\color{RoyalBlue}3 egg whites, room temperature}\\
Bob: {\color{magenta}whats room temperature.} \\
Alice: {\color{RoyalBlue}Room temperature is around 70 degrees Fahrenheit.}\\
Bob: {\color{magenta}what do you mean by medium bowl.} \\
Alice: {\color{RoyalBlue}A medium bowl is a bowl that is big enough to fit the ingredients for this recipe and that is comfortable to mix in.}\\
Bob: {\color{magenta}exactly how big should that be.} \\
Alice: {\color{RoyalBlue}(image) A medium bowl is a bowl that is big enough to fit the ingredients for this recipe and that is comfortable to mix in.}\\
{\color{Black}(a)} {\color{LimeGreen}video search {\color{Rose}\{recipe video\}} with "A medium bowl is a bowl that is..."}\\
Bob: {\color{magenta}ok assume im done whats next.} \\
Alice: {\color{RoyalBlue}Step: Keep beating and slowly add the sugar until stiff peaks form.}\\
Bob: {\color{magenta}whats stiff peaks.} \\
Alice: {\color{RoyalBlue}Stiff peaks are when the egg whites stand up in peaks that do not fall over.}\\
Bob: {\color{magenta}can you show me what that looks like.} \\
Alice: {\color{RoyalBlue}(image) Stiff peaks are when the egg whites stand up in peaks that do not fall over.}\\
{\color{Black}(b)} {\color{LimeGreen}video search {\color{Rose}\{recipe video\}} with "Stiff peaks are when the egg whites..."}\\
Bob: {\color{magenta}do i need to add water.} \\
Alice: {\color{RoyalBlue}No, you do not need to add water.}
        }
      }}
      \vspace{-0.5em}
    \end{flushleft}
  }}
  % \caption{VLM + LM prompt.}
  \label{fig:sub1}
\end{subfigure}%
\begin{subfigure}{.225\textwidth}
  \centering
  \vspace{1.0em}
  \includegraphics[width=1.0\linewidth]{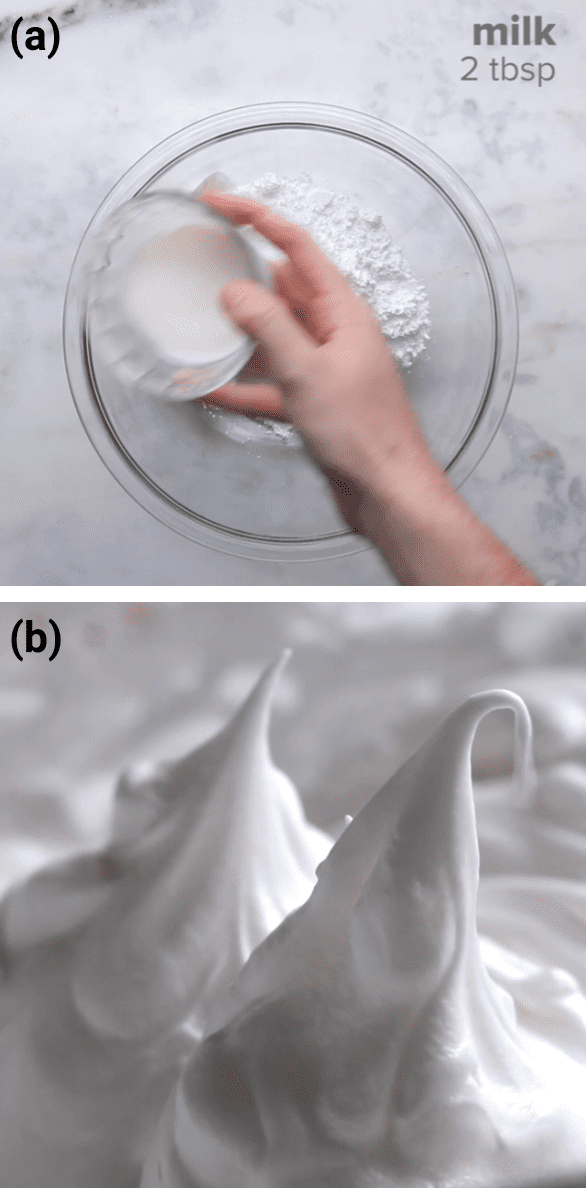}
  % \caption{Example results on MS COCO images.}
  \label{fig:sub2}
\end{subfigure}
\vspace{-1.5em}
\caption{\small SMs with {\color{LimeGreen}VLM}, {\color{Rose}Web Search}, and {\color{RoyalBlue}LM} prompting can enable multimodal dialogue applications such as guiding a user through online recipe steps and providing assistive visuals via video search.}
\label{fig:guided-recipes}
\end{figure}

\subsection{Robot Perception \& Planning: {\color{magenta}User} + {\color{LimeGreen}VLM} + {\color{RoyalBlue}LM} + {\color{Orange}Policies}}

\begin{figure}[H]
\centering
\begin{subfigure}{.61\textwidth}
  \centering
  \noindent\fbox{%
    \parbox{0.96\linewidth}{%
      \vspace{-0.5em}
      \begin{flushleft}\scriptsize{
        \texttt{{\color{Gray}
objects = [{\color{LimeGreen}"green block", "blue block", "yellow block", "green bowl", "blue bowl", "yellow bowl"}]\\
\# {\color{magenta}move all the blocks to different corners}.\\
Step 1. {\color{RoyalBlue}robot.pick\_and\_place("green block", "top left corner")}\\
Step 2. {\color{RoyalBlue}robot.pick\_and\_place("blue block", "top right corner")}\\
Step 3. {\color{RoyalBlue}robot.pick\_and\_place("yellow block", "bottom left corner")}\\
\# {\color{magenta}now move the blue block to the middle}.\\
Step 1. {\color{RoyalBlue}robot.pick\_and\_place("blue block", "middle")}\\
\# {\color{magenta}stack the blocks on top of each other}.\\
Step 1. {\color{RoyalBlue}robot.pick\_and\_place("yellow block", "blue block")}\\
Step 2. {\color{RoyalBlue}robot.pick\_and\_place("green block", "yellow block")}\\
\# {\color{magenta}wait actually undo that last step}.\\
Step 1. {\color{RoyalBlue}robot.pick\_and\_place("green block", "top left corner")}\\
\# {\color{magenta}put the yellow block in the bowl you think it best fits}.\\
Step 1. {\color{RoyalBlue}robot.pick\_and\_place("yellow block", "yellow bowl")}\\
\# {\color{magenta}ok now sort the remaining blocks in the same way}.\\
Step 1. {\color{RoyalBlue}robot.pick\_and\_place("blue block", "blue bowl")}\\
Step 2. {\color{RoyalBlue}robot.pick\_and\_place("green block", "green bowl")}
        }
      }}
      \vspace{-0.5em}
    \end{flushleft}
  }}
  % \caption{VLM + LM prompt.}
  \label{fig:sub1}
\end{subfigure}%
\begin{subfigure}{.39\textwidth}
  \centering
  \vspace{1.0em}
  \includegraphics[width=1.0\linewidth]{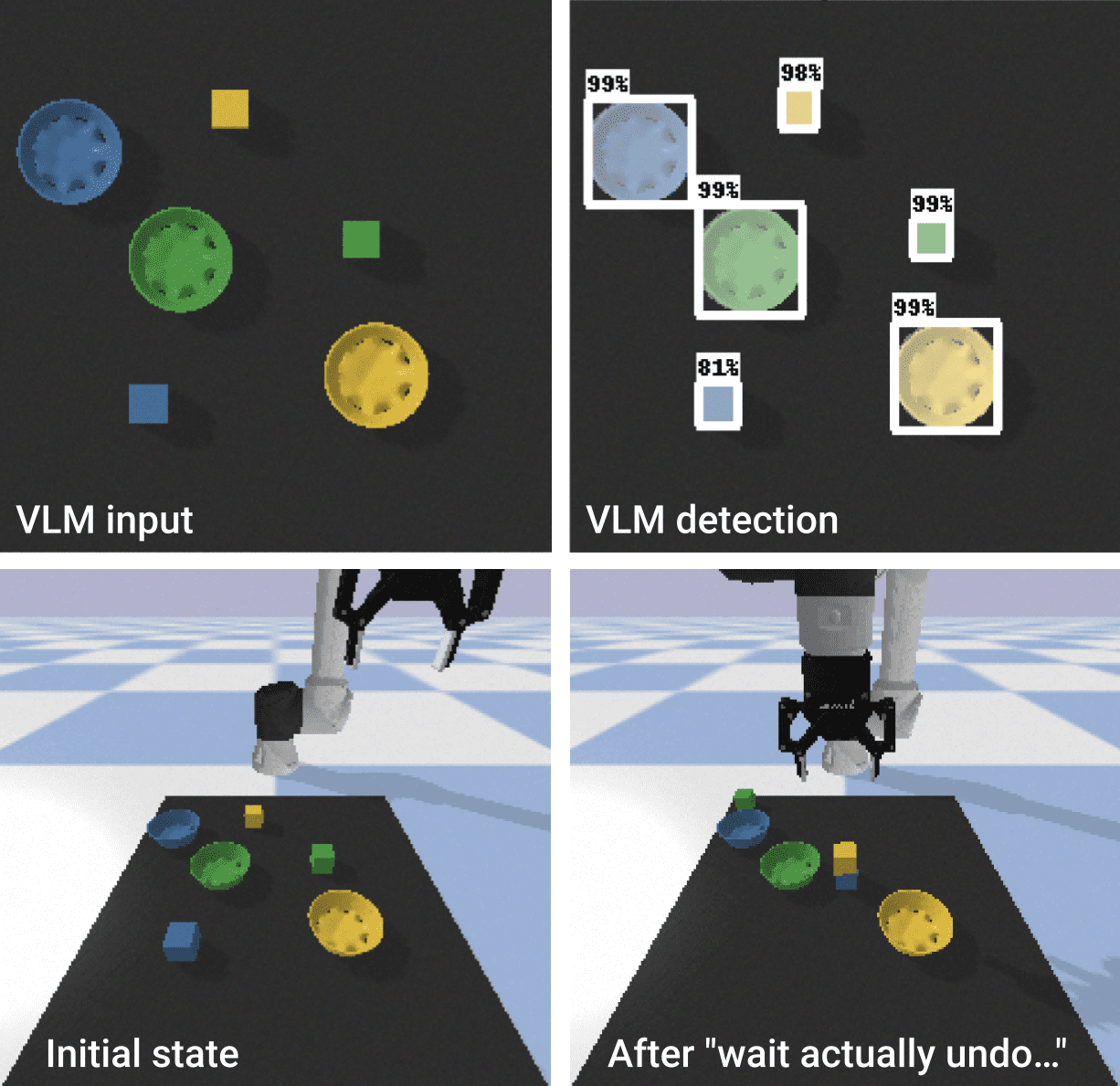}
  % \caption{Example results on MS COCO images.}
  \label{fig:sub2}
\end{subfigure}
\vspace{-1.5em}
\caption{\small SMs can be engineered with {\color{LimeGreen}VLM}, {\color{RoyalBlue}LM}, and language-conditioned {\color{Orange}robot policies} (\eg via CLIPort \cite{shridhar2022cliport}) to enable robots to parse and generate plans from free-form human instructions (in {\color{magenta}magenta}).}
\label{fig:robot-planning}
\end{figure}

SMs can be used to enable robots to perform language-conditioned tasks.
Our example system uses a VLM (open-vocabulary object detection with ViLD \cite{gu2021open}) to describe the objects in the scene, feeds that description as context to a LM as a multi-step planner \cite{saycan2022arxiv,huang2022language}, that then generates the individual steps to be passed to a pretrained language-conditioned robot policy (\eg models similar to CLIPort \cite{shridhar2022cliport,zeng2020transporter} for open vocabulary pick-and-place). Steps can be represented in the form of natural language (``{\color{RoyalBlue}Pick the red block and place it on the blue block.}'') or in the form of pseudocode (to generate text with a fixed template \eg ``{{\color{RoyalBlue}robot.pick\_and\_place("red block", "blue block")}''}), leveraging LM capacity to write code.
We demonstrate this in the context of a simulated environment (shown in Fig. \ref{fig:robot-planning}) using a UR5 arm and and several objects (blocks, bowls). Distinct from \cite{saycan2022arxiv}, this uses cross-modal VLM-informed prompting and LM code generation, rather than joint probabilistic inference.
%Example exchange with a user and LM generated plans:
\vspace{-2.0em}
\vspace{-1.0em}
Chaining this system together expands the set of language-specified tasks beyond the original set of primitives trained by the policy, and enables applications involving human dialogue with the robot.

\section{Discussion} \label{sec:discussion}

Socratic Models is a modular framework that leverages structured dialogue (\ie via prompting) between multiple large pretrained models to make joint predictions for new multimodal tasks. SMs leverage the commonsense knowledge already stored within foundation models pretrained on different domains of data (\eg text-to-text, text-to-images, text-to-audio), which may include Internet-scale data. Our shown systems for image captioning, video-to-text retrieval, egocentric perception, multimodal dialogue, robot perception and planning are just examples of the SMs framework, and may shed light on new opportunities to build simple systems that adapt pre-existing foundation models to (i) capture new multimodal functionalities zero-shot without having to rely on additional domain-specific data collection or model finetuning, and (ii) do so while retaining their robustness to distribution shifts (which is known to deteriorate after finetuning) \cite{wortsman2021robust}.  Potential future work may involve meta-learning the Socratic interactions themselves, and extending the inter-module edges to include additional modalities beyond language, \eg passing images between modules.

\textbf{Broader Impacts.} SMs offer new perspectives that encourage building AI systems using off-the-shelf large pretrained models without additional data collection or model finetuning.  
This leads to several practical benefits, new applications, and risks as well. For one, SMs provide an interpretable window, through language, into the behavior of the systems (even for non-experts). Further, the barrier of entry for this technology is small: SMs can be engineered to capture new functionalities with minimal compute resources, and to tackle applications that have traditionally been data-scarce. No model training was used to create our demonstrated results. This can be enabling, but also raises potential risks, since it increases the flexibility of unintended end use applications, and should be carefully monitored over time. It is also important to note that the system may generate results that reflect unwanted biases found in the Internet-scale data on which incorporated models are trained, and should be used with caution (and checked for correctness) in downstream applications.
We welcome broad discussion on how to maximize the potential positive impacts (enabling broad, new multimodal applications, with minimal new resources) while minimizing the capabilities of bad actors.
%\swelker one big impact that i see from SM is the world-state history becoming "readable, explainable and understandable" to humans because it is language. This enables future AI systems to become more transparent by using mutually compatible representations.

% \andy{move this to limitations:} It is also important to note that the system may generate captions that reflect unwanted biases found in the Internet-scale data on which incorporated models are trained on, and should be used with caution (and checked for correctness) in downstream applications.

% We are excited about opportunities in downstream applications as well. For example, SMs suggest promising research directions for data-driven learning in robotics, where various modules (\eg planning \cite{saycan2022arxiv,huang2022language}, perception \cite{shridhar2022cliport}) can be replaced with zero-shot foundation models imbued with commonsense priors across domains. These ideas may give rise to a new class of systems where by grounding affordances \cite{zeng2019learning} on language, control algorithms can begin to tap into the capabilities of models trained on Internet-scale data, to tackle applications that have traditionally been data-scarce.
%\swelker I think especially the control algorithms deserve follow up later, as we already saw a model correctly perform chat based bang-bang control after being explained the system.(bike) (this seems almost like the know how a human has to put into policy gradient type algorithms.)

\clearpage

\begin{ack}
We thank Debidatta Dwibedi, Matthew O'Kelly, and Kevin Zakka for excellent feedback on improving this manuscript, Anelia Angelova, Jean-Jacques Slotine, Jonathan Tompson, Shuran Song, for fruitful technical discussions, Kan Huang for applications support, Ahmed Omran, Aren Jensen, Malcolm Slaney, Karolis Misiunas for advice on audio models, and Cody Wanner for YouTube videos.
\end{ack}
\bibliographystyle{corlabbrvnat}
\small
\bibliography{references}
\normalsize

% Comment out these next handful of lines on/off for appendix
\clearpage
\appendix
\leavevmode\\
\begin{center}
{\bf{\LARGE Appendix for {\em Socratic Models}}}
\end{center}
\vspace{2.0em}
\input{appendix_source.tex}

\end{document}

%% file: appendix_source.tex
\section{Overview}

The appendix includes: (i) unsupervised evaluation for model selection, (ii) additional notes on main experiments, (iii) more details on applications to egocentric perception, (iv) scaling video search for language-based world-state history, (v) more details on robot perception and planning experiments, (vi) additional discussion on future work (\eg SMs for deductive reasoning) and (vii) broader impacts (\eg energy and resource consumption). For code, see %\mbox{\href{https://sites.google.com/view/socraticmodels}{sites.google.com/view/socraticmodels}}.
\mbox{\href{https://socraticmodels.github.io/}{socraticmodels.github.io}}.

\section{Unsupervised Socratic Model Selection}

The combination of complementary models, in which one may compensate for the weaknesses of the other, opens an interesting avenue for unsupervised evaluation of model performance. Since our metric of interest is the {\it combined} performance of \eg a VLM and a LM -- rather than asking the question: `(A): how well does this VLM perform in absolute?' for SMs, we can instead ask: `(B): how well does this VLM compensate for the weakness of the LM?'.

Strope et al. \cite{strope2011unsupervised} proposes a scheme which does so without requiring any evaluation ground truth. They also find that asking question (B) correlates well with answers to question (A), and is useful \eg for model selection. The method assumes you have access to a weak (wLM) and a strong (sLM) LM (respectively VLM if evaluating the LM's performance). Asking ``how well does this VLM compensate for the weaknesses of the LM'' is equivalent to asking: ``if we have a collection of VLMs, and we combine them with a weak LM, which model is going to perform the closest to the combination of the VLM with a strong LM?'' If a VLM combined with a weak LM, instead of a strong one, makes up for the LM's shortcomings and still performs well in combination, then it may serve as a better component in the context of this combined system.

The benefit of this approach -- while not entirely making up for doing absolute evaluations against a ground truth -- is that because it only measures relative distance between model outputs, it can be performed unsupervised without annotated data: the distance between the output of the weak and strong combination can be measured using measures of semantic distance, for instance here by scoring them against a distinct, held-out language model.

% The ideas behind SMs are relevant to classic work in speech recognition, where a traditional approach is to combine two complementary sources of semantic information with two models: an acoustic model and a language model. \cite{strope2011unsupervised} demonstrates it is possible to evaluate the individual performance of these models entirely unsupervised, without any labeled ground truth data. While the original motivation behind their work is to address the difficulties of collecting data and evaluating human transcription, similar concerns can be raised for multimodal domains such as egocentric video Q\&A, where ground truth labels are difficult to collect (and the quality of which can be highly subjective). The methods presented in \cite{strope2011unsupervised} provide practical benefits, such as fast characterization of prototype variations to help prioritize efforts in industrial production.

\begin{wraptable}{r}{0.53\textwidth}
%\begin{table}[h]
  \vspace{-1.0em}
  \setlength\tabcolsep{2.3pt}
  \centering
  \scriptsize
  \begin{tabular}{@{}lccccc@{}}
  \toprule
  & \multicolumn{5}{c}{VLM (CLIP) Variants + Weak LM}  \\
  \cmidrule(lr){2-6}
  Truth Models        & RN50x4 & RN50x16 & ViT-B/32 & ViT-B/16 &  ViT-L/14                      \\
  \midrule
  GPT-3 + ViT-B/16          & 0.628 & 0.646 & 0.686 & {\color{Gray}0.861} & \textbf{0.704} \\
  GPT-3 + RN50x16           & 0.667 & {\color{Gray}0.851} & 0.689 & 0.655 & \textbf{0.704} \\
  \midrule
  ImageNet Accuracy       & 65.8 & 70.5 & 63.2 & 68.6 & 76.2  \\
  Size (\# params)        & 178M & 291M & 151M & 150M & 427M  \\
  \bottomrule
  \end{tabular}
  \vspace{0.5em}
  \caption{\small{Unsupervised evaluation (higher is better) of various VLMs by pairing them with a weak LM and comparing outputs to a VLM paired with a strong LM, which provides relative `truth gradients' that inform how well the VLMs can compensate for the weak LM. These results suggest that better VLMs (measured by zero-shot ImageNet classification accuracies) can improve Socratic synergies. 
  }} % task, success \% vs. $\#$ of demonstrations. 
  \vspace{-1.0em}
  \label{table:results}
%\end{table}
\end{wraptable}

As an example of using this approach, we extend the method in Strope et al. \cite{strope2011unsupervised} to Socratic Models on egocentric perception, where we show it is possible to quantify the mutual dependence between foundation models without ground truth data. Specifically, to evaluate a new VLM (\textit{\textcolor{LimeGreen}{VLM\textbf{'}}}) for generating language-based world-state history, we first use a baseline VLM \textit{\textcolor{LimeGreen}{VLM}} paired with the strong LM (\textit{\textcolor{RoyalBlue}{\textbf{s}LM}}) to generate pseudo ground truth predictions \textit{\textcolor{LimeGreen}{VLM}}$\cross$\textit{\textcolor{RoyalBlue}{\textbf{s}LM}}. We then take both the baseline VLM \textit{\textcolor{LimeGreen}{VLM}} and new VLM \textit{\textcolor{LimeGreen}{VLM\textbf{'}}}, and pair them with a weak LM \textit{\textcolor{RoyalBlue}{\textbf{w}LM}} to generate predictions \textit{\textcolor{LimeGreen}{VLM}}$\cross$ \textit{\textcolor{RoyalBlue}{\textbf{w}LM}} and \textit{\textcolor{LimeGreen}{VLM\textbf{'}}}$\cross$\textit{\textcolor{RoyalBlue}{\textbf{w}LM}} respectively. We score these predictions (per image summary) against the pseudo ground truth \textit{\textcolor{LimeGreen}{VLM}}$\cross$\textit{\textcolor{RoyalBlue}{\textbf{s}LM}}. Since the outputs are linguistic, we can measure the similarity of a given prediction to the ground truth, by comparing their sentence embeddings produced by another language model \eg RoBERTa \cite{liu2019roberta}. It is important to use a distinct LM for scoring to avoid spurious correlations with the models under evaluation.

Tab. \ref{table:results} shows example results of this analysis with GPT-3 ``Davinci'' as the \textit{\textcolor{RoyalBlue}{\textbf{s}LM}}, and GPT-3 ``Curie'' as the \textit{\textcolor{RoyalBlue}{\textbf{w}LM}}, to compare VLM (\ie CLIP) variants with different backbones: vision transformers (ViT) \cite{dosovitskiy2020image} and ResNets (RN50) \cite{he2016deep} with different model sizes. We find that this method can capture a correlation of ascending performance curve with increasingly better VLMs (\eg better variants of CLIP) \cite{radford2021learning}, as measured by zero-shot image classification accuracy on ImageNet \cite{deng2009imagenet} -- with correlation coefficients of 0.41 and 0.46 between ImageNet accuracies and mean similarity to truth models via ViT-B/16 and RN50x16 respectively. We find that with our SM system for egocentric perception (and in contrast to the original setting in \cite{strope2011unsupervised}), it is necessary to use a third baseline VLM \textit{\textcolor{LimeGreen}{\textbf{b}VLM}}$\cross$\textit{\textcolor{RoyalBlue}{\textbf{s}LM}} to generate the pseudo ground truth, instead of \textit{\textcolor{LimeGreen}{VLM}}$\cross$\textit{\textcolor{RoyalBlue}{\textbf{s}LM}}. This is because the SM combinations that use the same VLM as the one that generates ground truth are biased to produce similar visual grounding results and can exhibit an unfair advantage during the comparisons. Those numbers in our tests have been {\color{Gray}grayed} out in Tab. \ref{table:results}.

\section{Additional Notes on Experiments}

\textbf{Choice of models.} There are many options of large pretrained ``foundation'' \cite{bommasani2021opportunities} models to choose from, but our experiments in the main paper use models that are publicly available, so that our systems can be made accessible to the community. In particular, we use CLIP \cite{radford2021learning} as the text-image similarity VLM (ViT-L/14 with 428M params, except on MSR-VTT which uses ViT-B/32), ViLD \cite{gu2021open} as the open-vocabulary object detector VLM; Wav2CLIP \cite{wu2021wav2clip} as the sound-critic ALM and Google Cloud Speech-to-text API \cite{gcloud-speech-to-text} as the speech-to-text ALM; GPT-3 with 175B params \cite{brown2020language,ouyang2022training} and RoBERTa \cite{liu2019roberta} with 355M params as the LMs. 
All pretrained models are used off-the-shelf with no additional finetuning. In terms of compute resources required, all experiments can be run on a single machine using an NVIDIA V100 GPU with internet access for outsourced API calls (\eg GPT-3 and Google Cloud Speech-to-text).

\subsection{Image Captioning on MS COCO}

\begin{wraptable}{r}{0.53\textwidth}
% \begin{table*}[h]
  \vspace{-3.15em}
  \setlength\tabcolsep{2.0pt}
  \scriptsize
  \begin{tabular}{@{}lcccccc@{}}
  \toprule
  Method         &  BLEU-4 & METEOR &  CIDEr & SPICE & ROUGE-L \\
  \midrule
  $^*$ClipCap \cite{mokady2021clipcap} (full) & 33.5 & 27.5 & 113.1 & 21.1 & --\\
  $^\dagger$MAGIC \cite{su2022language} (full) & 12.9 & 17.4 & 49.3 & 11.3 & 39.9\\
  ZeroCap \cite{tewel2021zero} (full) & 2.6 & 11.5 & 14.6 & 5.5 & --\\
  \midrule
  $^*$ClipCap \cite{mokady2021clipcap} (subset) & 40.7 & 30.4 & 152.4 & 25.2 & 60.9\\
  $^\dagger$MAGIC \cite{su2022language} (subset) & 11.4 & 16.4 & 56.2 & 11.3 & 39.0\\
  ZeroCap \cite{tewel2021zero} (subset) & 0.0 & 8.8 & 18.0 & 5.6 & 18.3\\
  \bottomrule
  \end{tabular}
  
  \vspace{0.3em}
  $^*$finetuned on full training set with image-text pairs.\\
  $^\dagger$finetuned on unpaired training set, zero-shot on image-text pairs.
  \vspace{-0.5em}
  \caption{\small{Image captioning metrics on the random subset of $N=100$ (bottom) test examples are comparable to the full MS COCO test set metrics (top).
  }}
  \vspace{-2.0em}
  \label{table:mscoco-appendix}
  % table: https://docs.google.com/document/d/1nPJH1DkaBN7vqX12b3bxAszbW-YyJm4AEXrDmaZuig0/edit#
% \end{table*}
\end{wraptable}

For image captioning experiments on the MS COCO dataset \cite{chen2015microsoft,lin2014microsoft}, we evaluate over a random sampled subset of 100 images from the test split \cite{karpathy2015deep}, so that GPT-3 API runtime costs are more affordable for reproducibility ($\sim$\$150 USD per run with with $n=20$ generated candidate captions per image). Metrics (shown in Tab. \ref{table:mscoco-appendix}) from baselines reported on this subset of MS COCO test examples are comparable to the full test set metrics. Also, while the captions in Fig.~3, Section 4.1, were generated with the prompt ``\ldots creative short\ldots'' as noted in Fig.~3, for best quantitative MS-COCO captions we used the prompt ``\ldots short, likely\ldots''.

\subsection{Contextual Image Captioning on Concadia}

Our experiments on Concadia \cite{kreiss2021concadia} evaluate the extent to which SMs can generate captions and descriptions conditioned on input images and their associated article text. While our results show that the SM combination of VLMs and LMs can achieve strong results on the benchmark, we also observe that LMs (\eg GPT-3) alone can return surprisingly competitive results too (Tab.~\ref{table:concadia-results-full}). Specifically, using the same LM prompt from the SM approach, but leaving out information from the VLM:

\noindent\fbox{%
    \parbox{0.97\linewidth}{%
        \scriptsize{\texttt{{\color{Gray}I am an intelligent image captioning bot. The article is about: "{\color{Black}\{article\_text\}}". \sout{In this image, I think I see a {\color{LimeGreen}\{object1\}}, {\color{LimeGreen}\{object2\}}, {\color{LimeGreen}\{object3\}},...} A short caption for this image is:}}}
    }%
}

\begin{wraptable}{r}{0.47\textwidth}
  \vspace{-1.0em}
  \setlength\tabcolsep{2.0pt}
  \scriptsize
  \begin{tabular}{@{}lccc@{}}
  \toprule
  Method         &  Caption Generation & Description Generation \\
  \midrule
  Kreiss et al. \cite{kreiss2021concadia} & 11.3 & 17.4\\
  SMs (ours) & 38.9 & 22.6\\
  \textbf{SMs (no image)} & 40.1 & 20.6\\
  {\color{Gray}SMs w/ description} & {\color{Gray}93.8} & {\color{Gray}--}\\
  \bottomrule
  \end{tabular}
  \caption{\small{SMs on zero-shot contextual image captioning and description tasks on the Concadia dataset.
  }}
  \vspace{-3.0em}
  \label{table:concadia-results-full}
\end{wraptable}

subsequently drops performance on image description by 2.0 CIDEr points, but surprisingly improves captioning performance by 1.2 CIDEr points. This suggests: (i) information from the VLM is more important for LMs in generating image descriptions than captions, (ii) there may be a strong correlation between the distributions of captions and article texts that can be leveraged by an LM alone, and/or (iii) there may exist overlap between Concadia (\eg Wikipedia articles) and the training set of the LM, which warrants further investigation to disentangle confounding variables.

\subsection{Video-to-text Retrieval on MSR-VTT 1k-A}

We also report results in Tab. \ref{table:msr-vtt-appendix} on the popular MSR-VTT ``1k-A'' subset, introduced by Yu et al. \cite{yu2018joint} created via random sampling on the full test set. We follow the same evaluation protocol for video-to-text retrieval as used in prior work \cite{liu2019use,dong2019dual,dong2018predicting,mithun2018learning}, which reports the minimum rank among all valid text captions for a given video query, and each test video is associated with 20 captions.

\begin{table}[H]
\setlength\tabcolsep{2.8pt}
  \centering
  \scriptsize
  \begin{tabular}{@{}llccccccccll@{}}
  \toprule
  \multicolumn{2}{l}{} & \multicolumn{4}{c}{MSR-VTT 1k-A} \\
%   \multicolumn{2}{l}{} & \multicolumn{3}{c}{Text-to-Video Retrieval} & \multicolumn{3}{c}{Video-to-Text Retrieval} &  &  \\
  \cmidrule(lr){3-6}
  Category & Method         & R@1$\uparrow$ & R@5$\uparrow$ & R@10$\uparrow$ & MdR$\downarrow$    & Audio & CLIP enc.        \\
  \midrule
  \multirow{8}{*}{\em{Finetuning}} & Collaborative Experts \cite{liu2019use} & 20.6 & 50.3 & 64.0 & 5.3 & yes &  \\
                                   & SSB \cite{patrick2020support}  & 28.5 & 58.6 & 71.6 & 3.0 & no &  \\
                                   & CLIP4Clip \cite{luo2021clip4clip}  & 43.1 & 70.5 & 81.2 & 2.0 & no & ViT-V/32 \\
                                   & CLIP2Video \cite{fang2021clip2video}  & 43.5 & 72.3 & 82.1 & 2.0 & no & ViT-V/32 \\
                                   & DRL \cite{wang2022disentangled}, ViT-B/32  & 45.3 & 73.9 & 83.3 & 2.0 & no & ViT-V/32 \\
                                   & CAMoE \cite{cheng2021improving} & 49.1 & 74.3 & 84.3 & 2.0 & no & ViT-B/32 \\
                                   & CLIP2TV \cite{gao2021clip2tv}     & 54.1 & 77.4 & 85.7 & \textbf{1.0} & no & ViT-B/16 \\
                                   & DRL \cite{wang2022disentangled}, ViT-B/16 + QB-n   & \textbf{56.2} & \textbf{79.9} & \textbf{87.4} & \textbf{1.0} & no & ViT-B/16 \\
  \midrule
  \multirow{3}{*}{\em{Zero-shot}} & SSB \cite{patrick2020support}, zero-shot & 8.7 & 23.0 & 31.1 & 31.0 & no &  \\      
                                %   & CLIP via \cite{portillo2021straightforward} & 27.2 & 51.7 & 62.6 & 5.0 & no & ViT-B/32 \\
                                  & CLIP via \cite{portillo2021straightforward} & 58.0 & 82.5 & 90.2 & 1.0 & no & ViT-B/32 \\
                                   & SMs (ours)   & \textbf{60.7} & \textbf{84.1} & \textbf{90.6}  & \textbf{1.0} & yes & ViT-B/32 \\
  \bottomrule
  \end{tabular}
  \vspace{0.5em}
  \caption{\small{Video-to-text retrieval results on MSR-VTT \cite{xu2016msr} dataset on the 1k-A \cite{yu2018joint} subset. Differentiated are methods which train on the MSR-VTT dataset ({\em{finetuning}}), compared with {\em{zero-shot}} methods, which do not. Also noted: whether the methods use audio channels, and if CLIP \cite{radford2021learning} is used, which CLIP encoder is used.
  }}
  \vspace{-2.0em}
  \label{table:msr-vtt-appendix}
  % table: https://docs.google.com/document/d/1nPJH1DkaBN7vqX12b3bxAszbW-YyJm4AEXrDmaZuig0/edit#
\end{table}

Note that the original CLIP baseline for video-to-text retrieval via Portillo-Quintero et al. \cite{portillo2021straightforward} reports R@1 to be 27.2, but this was computed with only 1 caption per video that was random sampled \cite{yu2018joint} from the original set of 20 captions (for text-to-video retrieval). This differs from the original evaluation protocol and may be sub-optimal since the sampled caption can be ambiguous or partial (generated from crowd compute). For example, videos may be paired with a vague caption ``a person is explaining something'' as ground truth, rather than one of the other (more precise) captions \eg ``a person is talking about importing music to a ipod''. Upon correcting the evaluation protocol (\ie increasing the number of associated captions per video to 20), R@1 for Portillo-Quintero et al. \cite{portillo2021straightforward} improves to 58.0, and SMs improve on top of that with LMs and ALMs\footnote{\scriptsize Key used parameters for Google Cloud Speech-to-Text API include `\texttt{model=video}' and `\texttt{use\_enhanced=True}'. At  0.006 cents per 15 seconds, this represents an estimated speech-to-text processing cost of under 25 cents (USD) for all MSR-VTT test data.} to 60.7 R@1 zero-shot.

% We may also prompt LMs (via multiple-choice) to determine if one caption is a better fit than another for a given video. However, for this specific task and dataset, with thousands of possible answers to choose from, the numerical ranking provided by embedding similarity scores provides a practical solution rather than relying on thousand-way multiple-choice commonsense reasoning.

% Like other recent works \cite{gao2021clip2tv}, we focus our results on this dataset. One of the reasons this is a good task and dataset for generally testing the value of the SMs approach is that there is already a strong zero-shot baseline, provided by Portillo-Quintero et al. \cite{portillo2021straightforward}, which uses CLIP by itself, but does not use the Socratic method: there is no multi-model exchange, and no LMs are used. Additionally, this task provides a great opportunity to incorporate another type of modality -- speech-to-text from audio data.

% Note that instead of {\em{video-to-text}} retrieval, but rather on {\em{text-to-video}} retrieval, a recent method \cite{li2022blip} has shown strong zero-shot results.
Other methods have also evaluated on zero-shot MSR-VTT {\em{text-to-video}} retrieval \cite{xu2021videoclip,miech2020end,bain2021frozen}, but these have all been outperformed by Portillo-Quintero et al. \cite{portillo2021straightforward}. Our method may be adapted as well to {\em{text-to-video}}, but due to our use of transcripts on only a subset of the videos, unlike in video-to-text, this creates an asymmetry which may require an unwieldy relative weighting for ranking videos with or without transcripts. Note that (Tab.~\ref{table:msr-vtt-appendix}) prior to the CLIP revolution in video-to-text retrieval, using the audio modality was not uncommon amongst competitive video-to-text retrieval methods \cite{mithun2018learning,liu2019use}. The trend over the past year, however, has been to instead focus on using only visual features, with {\em{all}} recent competitive methods being based off of CLIP, and not using audio data. Our approach, through leveraging commonsense reasoning stored in the LMs, is able to once again allow audio data to enable progress in this common video understanding task, beyond what CLIP alone can provide.

\section{Egocentric Perception Appendix}

\textbf{Background.} Egocentric perception continues to be an important problem in computer vision.
Early work in the area explores hand-designed first-person visual features for egocentric action recognition, object understanding, and video summarization. This includes ego-motion (\eg optical flows) \cite{kitani2011fast,ryoo2013first} as well as features from human gaze, hands, and objects \cite{spriggs2009temporal,lee2012discovering,fathi2011understanding,pirsiavash2012detecting,li2013pixel,lee2015predicting}. Focusing on hand-designed features was common in early egocentric vision research, as the availability of data (or videos in general) was very limited.
%\michael{Still working on the below.}
%Early work in this area commonly explores either hand-designed first-person visual features \cite{kitani2011fast,fathi2011understanding,ryoo2013first,li2013pixel,lee2015predicting} or learned representations \cite{ma2016going}, but the lack of sufficient egocentric data consistently remains a bottleneck. 
More recent approaches in egocentric perception leverage learned feature representations, utilizing pretrained convolutional network features \cite{ryoo2015pooled}, finetuning them \cite{ma2016going,zellers2022merlot}, or training them from scratch \cite{bambach2015lending} with first-person videos. Similar to the topics explored in early work, learning of visual representations capturing human hands, objects, and eye gaze has been extensively studied \cite{garcia2018first,li2018eye}. \cite{kazakos2019epic} learns multimodal embeddings (\ie video + audio), and \cite{furnari2019would} studies future action anticipation from egocentric videos. Lack of sufficient data however, consistently remains a bottleneck -- motivating researchers to construct new larger-scale egocentric video datasets including EPIC-Kitchens \cite{damen2018scaling}, Charades-Ego \cite{sigurdsson2018charades}, and Ego4D \cite{grauman2021ego4d}.

\begin{figure*}[t]
  \centering
  \vspace{0.5em}
  \includegraphics[width=0.99\linewidth]{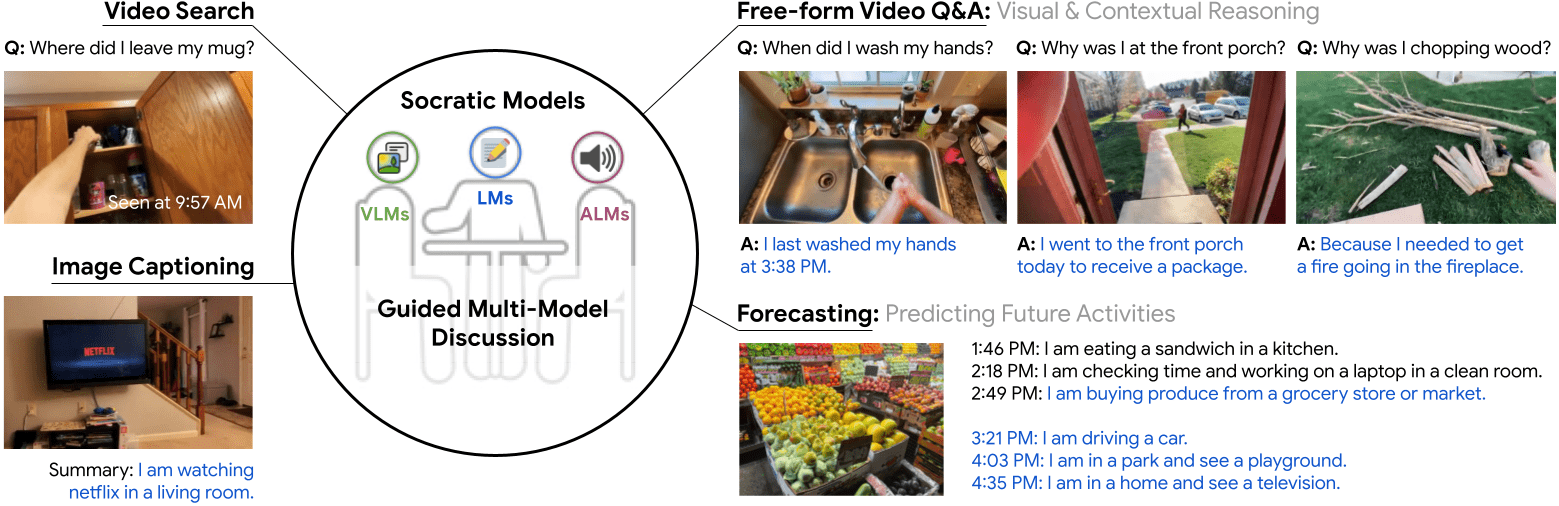}  
  \caption{\small{
  %SMs are a class of systems built on dialogue between pre-existing foundation models. While this paper presents a case study on egocentric perception, SMs can in practice be applied to other domains. In robotics, an interactive agent can be defined as an ensemble of VLMs, LLMs, and language-conditioned controllers (\eg indexing pretrained skills). Fairness can implemented as an additional node that guides the dialogue of two other models. Human-to-machine feedback loops can be added as language-based nodes that process user input to the system.
  % In this work we propose Socratic Models (SMs), a framework that uses structured dialogue between pre-existing foundation models, each of which can exhibit unique (but complementary) capabilities depending on the distributions of data on which they are trained.
  On various egocentric perceptual tasks (shown), this work presents a case study of SMs with visual language models ({\color{LimeGreen}VLMs, \eg CLIP}), large language models ({\color{RoyalBlue}LMs, \eg GPT-3, RoBERTa}), and audio language models ({\color{Purple} ALMs, \eg Wav2CLIP, Speech2Text}). From video search, to image captioning; from generating free-form answers to contextual reasoning questions, to forecasting future activities -- SMs can provide meaningful results for complex tasks across classically challenging computer vision domains, without any model finetuning.
  % \vikas{how is meaningfulness validated?}
  % \pete{Make so the whole black circle is SMs.  Maybe something like ``Socratic Models'' as a label above the circle, together with the guided multi-model dsicussion text?  Can then also try to make it so the three stick figures are respectively GPT-3, CLIP, and Wav2CLIP.}
  % Free-form visual question answering on egocentric videos can be performed using SMs without additional training. SMs (Model-to-Models) formulate perceptual cognition as dialogue between two large language-based models trained on the Internet: (i) a visual language model that grounds visual content on text, and (ii) a generative large language model for commonsense reasoning. In these examples, the system takes as input an egocentric video and a natural language question, then outputs a relevant frame, an answer, and an explanation.
  % https://docs.google.com/drawings/d/1XBaamqr86K3WYXw96ybo4R51dk9qZI65c6PnvzNt54Y/edit?usp=sharing
  % pete v3: https://docs.google.com/drawings/d/1-MeaOgnVDKdGRY-_0dlEmqPbf7O1LlouYQjqUSijPjs/edit
  }}
  \vspace{-1.0em}
  \label{fig:teaser}
\end{figure*}

\subsection{Why Egocentric Perception?}\label{subsec:why-egocenric}

We highlight SMs on egocentric perception because it is an important yet challenging computer vision domain \cite{grauman2021ego4d,damen2020rescaling,sigurdsson2018charades} with downstream applications in augmented reality (AR) and robotics \cite{saycan2022arxiv}. From unusual viewpoints to the lack of temporal curation -- the characteristics of first-person videos are unique and not often found in existing datasets, which focus more on generic Internet content captured from third-person spectator views \cite{deng2009imagenet,lin2014microsoft,sharma2018conceptual}. Notably, this domain shift makes it difficult for data-driven egocentric models to benefit from the standard paradigm of pretraining on third person Internet data \cite{li2021ego,sigurdsson2018charades}. % See Related Work (Sec.~\ref{sec:background-and-related}) for a more detailed discussion on prior work in egocentric perception --
Overall, the key challenges have included how to acquire sufficient egocentric data, and/or how to make sufficient use of this data (either with dense labels, or otherwise).

Despite the challenges of egocentric perception, we find that SMs can reconcile the complementary strengths of pretrained foundation models to address these difficulties through contextual reasoning.
For example, while modern activity recognition models trained on third person data might over-index to the motion of the primary person in video (making the models difficult to be adapted to first-person videos), we find that LMs like GPT-3 can suggest equally plausible activities (\eg ``receiving a package'') that may be occurring given only a brief description of the scene (\eg ``front porch'') and the objects detected in the image (``package, driveway, door'') by a VLM. 
These activity suggestions are often more expressive than the class categories that can be found in typical activity recognition datasets (\eg Charades \cite{sigurdsson2018charades}, Kinetics \cite{smaira2020short}), and reflect the information already stored in the models, agnostic to the point of view.
Our SM system for egocentric perception leverages these advantages, and also suggests future research directions in contextual reasoning that leverage existing language-based models without having to curate large annotated datasets.

\begin{figure*} %[t]
    \vspace{-1.5em}
    \captionsetup{type=figure}
    \includegraphics[width=0.96\textwidth]{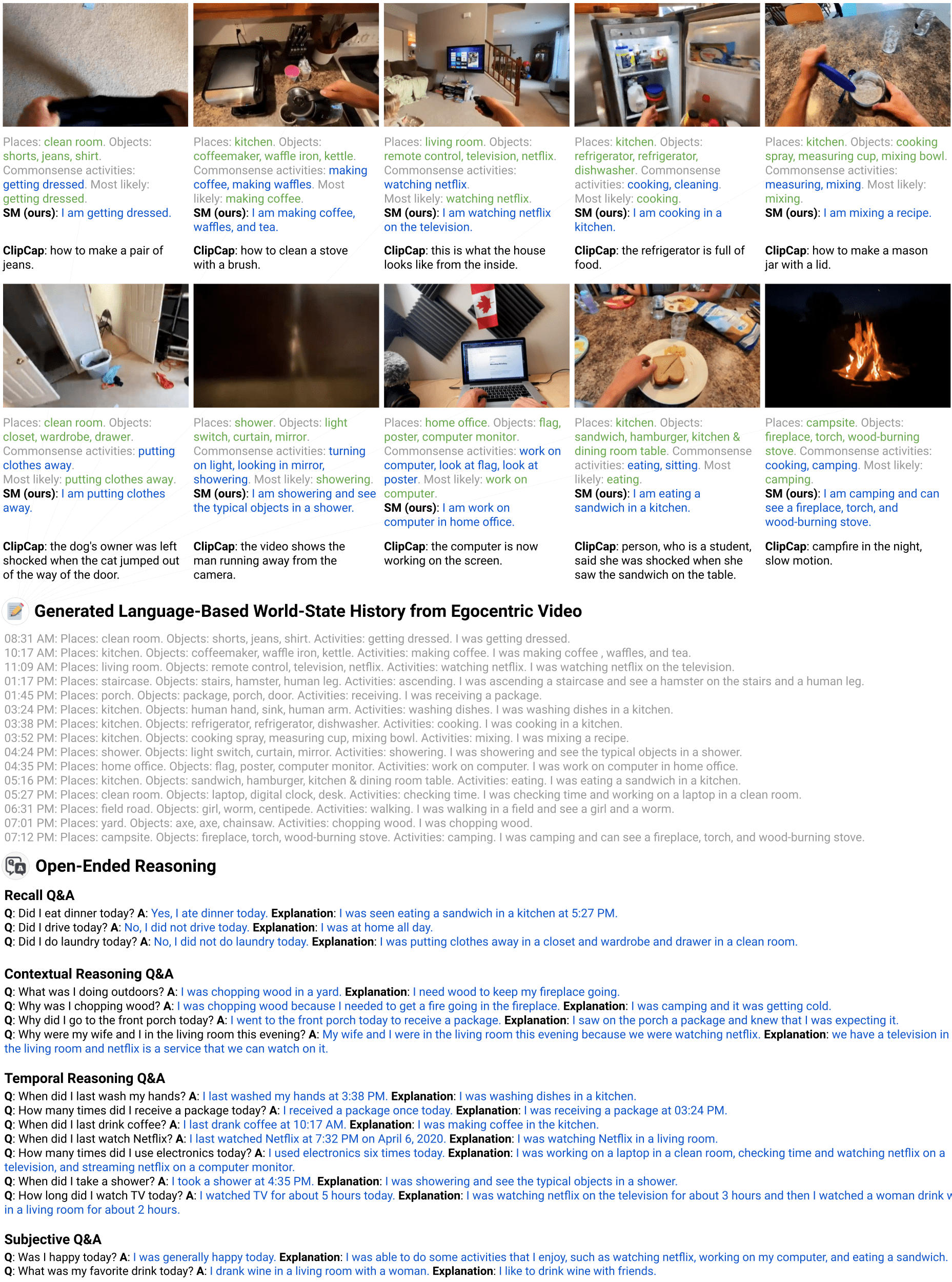}
    \captionof{figure}{\small An instantiation of the SMs framework for open-ended reasoning with egocentric perception. SMs can generate meaningful structured captions (top) for egocentric images through Socratic dialogue between VLMs ({\color{LimeGreen}green}) and LMs ({\color{RoyalBlue}blue}), and qualitatively perform well versus state-of-the-art captioning models such as ClipCap \cite{mokady2021clipcap}. Key moments from egocentric video are summarized with SMs into a language-based world-state history (middle), which can be provided as context to an LM for open-ended question answering. Results (bottom) for generated answers ({\color{RoyalBlue}blue}) and model explanations ({\color{RoyalBlue}blue}) suggest SMs are fairly capable of performing a variety of reasoning tasks including answering binary yes or no questions, contextual and temporal reasoning questions, as well as subjective questions.}
    \label{fig:results}
\end{figure*}
% v2 drawing: https://docs.google.com/drawings/d/11PQNMBHATKJs45vCp_dQlq2UvDOUml9lxIPM5SL2X94/edit?usp=sharing&resourcekey=0-rij4wpGMn6H3QSrrmblh8Q
% pete v3 drawing: https://docs.google.com/drawings/d/19yU7psBMcRXORz4nsrcVDJf5dqH_aUb0t5UvEXZYWfI/edit

\begin{figure*}[t]
  \centering
  \vspace{0.5em}
  \includegraphics[width=0.99\linewidth]{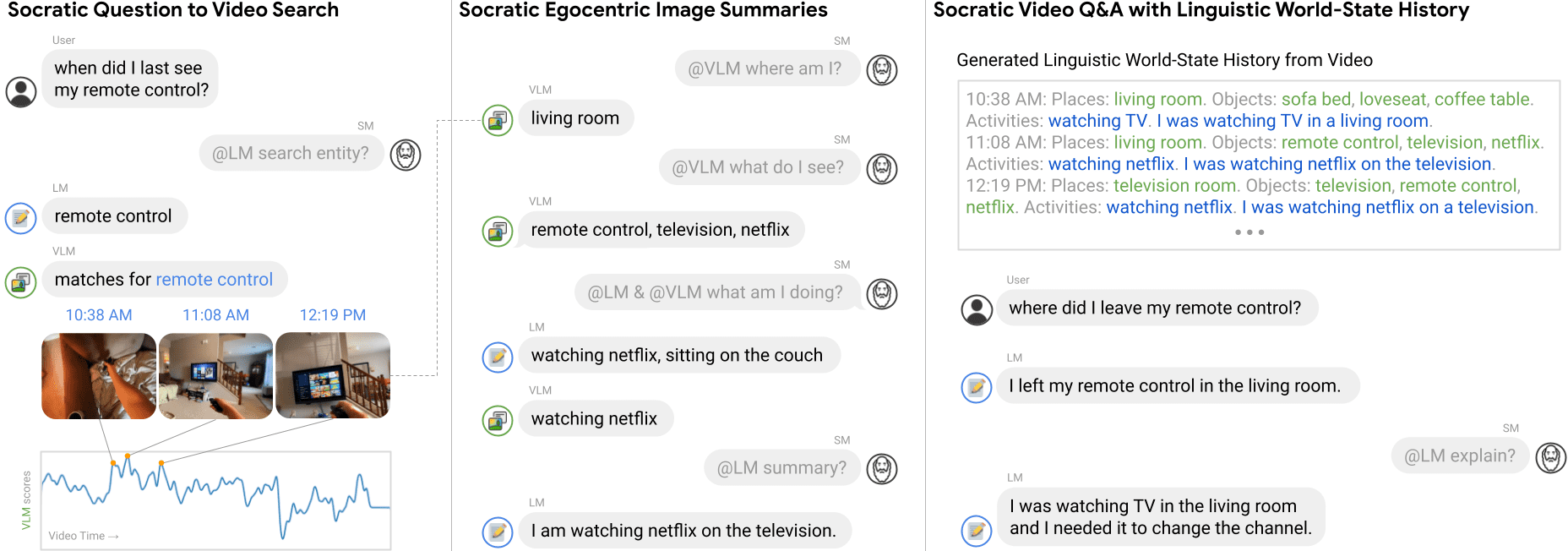}  
  \caption{\small{
  Examples of guided multi-model exchanges (Socratic Models) for an egocentric perception system%
  %-- shown above include a large language model (LM \eg BERT, GPT-3) and visual language model (VLM \eg CLIP, LiT): 
  % adding back (LM) and (VLM) to show which part is taken care of by which model
  : (i, left) parsing a natural language question %(LM \eg GPT-3)
  into search entities (with LM) to be used to find the most relevant key moments %(VLM \eg CLIP, LiT; or ALM \eg Wav2CLIP)
  in the video (with VLM); (ii, middle) describing each key frame by detecting places and objects (VLM), suggesting commonsense activities (LM), pruning the most likely activity (VLM), then generating a natural language summary (LM) of the SM interaction; (iii, right) concatenating key frame summaries into a language-based world-state history that an LM can use as context to answer the original question. %\vikas{1. "left" what is the distinction between "user" and "host". Should "host" be "human" for more clarity? Perhaps color "VLM", "LM" and "Human" tags differently?}
  % \pete{Use panel-level titles to more clearly call out what is happening.  Use ``language-based world-state history'' in figure and caption.}
  % https://docs.google.com/drawings/d/1hBbVJxkjB6mT1OR_a2ApMr2qr0v8A7hEWlmYXuzuUSE/edit
  }}
  \label{fig:system}
\end{figure*}

\subsection{Additional Details on Language-Based World-State History from Video}\label{subsec:linguistic-world-state}

In order to provide language-based reasoning capabilities for open-ended question-answering, a key aspect of our system is to describe the observed states of the world in language, with the goal of creating a language-based world-state history (Fig. \ref{fig:results}) that can be used as context to an LM. To this end, a component of our method generates Socratic image summaries of individual video frames (Sec. 3.3-A), that can then be concatenated (along with timestamps) to form an event log (illustrated at the top and middle of Fig.~\ref{fig:results}).

%swelker this also presents the opportunity to compress subsequent "same acitivity" or same objects visible into one "activity"

%The third component treats this event log as a short story, appends the original question, then prompts an LM to respond to it with a free-form answer.
%A key question, however, is how to determine which are the key frames?  We first discuss \textit{A.} Socratically forming single-frame image summaries, and then \textit{B.} how to assemble single-frame image summaries into a language-based world-state history usable for question-answering.

{\textbf{\textit{3.3-A. Socratic Egocentric Image Summaries.}}}
Given an image frame as input, this component generates a natural language summary (\eg caption) of what is occurring in the image.
Our system uses a Socratic approach with guided multimodal multi-model discussion to provide answers to 3 questions that describe the visual scene: ``where am I?'', ``what do I see?'', and ``what am I doing?'', which are then summarized into a single caption per image frame.

%\vikas{Fairly dense text below, split into subsections on "Place Identification", "Activity Recognition" etc?}
%

\begin{itemize}[leftmargin=*]
\item\textbf{Where am I?} For place recognition, we use a VLM to rank Places365 \cite{zhou2016places} scene categories against the image, with the top $n$ candidates (out of 365) inserted into a prefix: ``Places: {\color{LimeGreen}\{place1\}}, {\color{LimeGreen}\{place2\}}, {\color{LimeGreen}\{place3\}}.''.
\item {\textbf{What do I see?}} For object and people recognition, we use a VLM to rank OpenImages object categories \cite{kuznetsova2020open} against the image, with the top $m$ categories (out of 600) inserted into a second prefix: ``Objects: {\color{LimeGreen}\{object1\}}, {\color{LimeGreen}\{object2\}}, {\color{LimeGreen}\{object3\}}.''
\item {\textbf{What am I doing?}} For activity recognition, we use a back-and-forth interaction between an LM and VLM: we first use an LM to infer the activities most related to the places and objects previously listed by the VLM ({\color{LimeGreen}green}):

\noindent\fbox{%
    \parbox{0.97\linewidth}{\small{\texttt{{\color{Gray}Places: {\color{LimeGreen}\{place1\}}, {\color{LimeGreen}\{place2\}}, {\color{LimeGreen}\{place3\}}. Objects: {\color{LimeGreen}\{object1\}}, {\color{LimeGreen}\{object2\}}, {\color{LimeGreen}\{object3\}}. Activities:} {\textbf{\color{RoyalBlue}activity\_a, activity\_b, activity\_c.}}
    }}}}
% \noindent\fbox{%
%     \parbox{0.97\linewidth}{\small{\texttt{{\color{Gray}Places: {\color{LimeGreen}\{place1\}}, {\color{LimeGreen}\{place2\}}, {\color{LimeGreen}\{place3\}}. Objects: {\color{LimeGreen}\{object1\}}, {\color{LimeGreen}\{object2\}}, {\color{LimeGreen}\{object3\}}. Activities:} activity\_a, activity\_b, activity\_c.
%     }}}}

We find that generating candidate activities using an LM yields more suitable descriptions of egocentric activities and interactions with first-person video, than using standard activity recognition dataset categories (\eg from Charades or Kinetics). Activity recognition datasets are often tailored to third person videos, and can only cover a partial subset of human activities, which instead can be more holistically captured through LM reasoning \cite{petroni2019language} over the objects and places that the VLM perceives. For example, ``receiving a package'' is a common household activity not found in most datasets. After the LM generates candidate activities, these candidates are then fed back to the VLM and re-ranked to sort out the top $k$ activities by relevance to the key image frame: ``Activities: {\{\greenuline{activity1}\}}, {\{\greenuline{activity2}\}}, {\{\greenuline{activity3}\}}.'' 
\end{itemize}

This process of generating candidate activities from places and objects is one way of extracting commonsense from LMs as knowledge bases \cite{petroni2019language}. Continuing the Socratic dialogue further, this can be repeated likewise to generate new relevant objects (conditioned on activities and places), as well as new places (conditioned on objects and activities). 
One can iterate the procedure (LM generate, VLM re-rank, repeat) to populate the set of places, objects, and activities until equilibrium (\ie no more new entities), which generally helps to cover a broader set of places and objects that expand beyond the initial seed categories from Places365 and OpenImages. For example:  % \pete{If we want to involve this recursive discussion element, maybe should have a figure on this.} \johnny{this paragraph also seems out of place.  so, if there aren't compelling examples of this yet, perhaps put it in future rumination section near the end.} \andy{we do have a snippet in the code that can expand on the categories -- I can generate an example or two and drop them in here soon}

\noindent\fbox{%
    \parbox{0.97\linewidth}{\small{\texttt{{\color{Gray}If I am making {\greenuline{making pancakes}}, objects that I am likely to see include:} {\textbf{\color{RoyalBlue}a frying pan, a spatula, a bowl, milk, eggs, flour, sugar, baking powder, butter, a plate, syrup.}}
    }}}}
    
Given the final set of places, objects, and activities, we use the LM to generate an overall first-person summary of what is happening in the image.
Specifically, the prompt is:

\noindent\fbox{%
    \parbox{0.97\linewidth}{%
        \small{\texttt{\color{Gray}I am in a {\color{LimeGreen}place1}, {\color{LimeGreen}place2}, {\color{LimeGreen}place3}. I see a {\color{LimeGreen}object1}, {\color{LimeGreen}object2}, {\color{LimeGreen}object3}. I am {\greenuline{activity1}}. Question: What am I doing? Answer: I am most likely}}
    }%
}
% version with alternating colors
% \noindent\fbox{%
%     \parbox{0.97\linewidth}{%
%         \small{\texttt{\color{Gray}I am in a {\color{LimeGreen}place1}, {\color{LimeGreen}place2}, {\color{LimeGreen}place3}. I see a {\color{LimeGreen}object1}, {\color{LimeGreen}object2}, {\color{LimeGreen}object3}. I am {\color{RoyalBlue}acti\color{LimeGreen}vity1}. Question: What am I doing? Answer: I am most likely}}
%     }%
% }

%
The summarization process in general can capture more rich descriptions conditioned on the places, objects, and activities, and qualitatively seem to do well at ignoring irrelevant categories (\ie denoising). For example:% $\rightarrow$ (response) ``\texttt{}''.

\noindent\fbox{%
    \parbox{0.97\linewidth}{%
        \small{\texttt{{\color{Gray}I am in a {\color{LimeGreen}nursing home}, {\color{LimeGreen}landfill}, {\color{LimeGreen}living room}. I see a {\color{LimeGreen}wine}, {\color{LimeGreen}wine glass}, {\color{LimeGreen}woman}. I am {\greenuline{drinking wine}}. Question: What am I doing? Answer: I am most likely} {\textbf{\color{RoyalBlue}enjoying a glass of wine with a friend or loved one.}}}}
    }%
}\\

However, while the LM's denoising capabilities can compensate for the shortcomings of the VLM, it is important to note that this may also cause unwanted ignoring of notable, but rare events (\eg such as witnessing a purple unicorn, which may be ignored, but potentially it is Halloween). Finding new ways in which such events can be indexed appropriately may be useful for downstream applications.

% \pete{I think an important thing to point out is that the LM denoising capabilities we see might also cause unwanted ignoring of notable, but rare events.  We can make a simple example of this by creating an event log with "saw a purple unicorn", and then ask for a summary of the day.  The LM probably wouldn't include the purple unicorn sighting in there, but certainly if a human saw a purple unicorn, it would make their list of ``summary for the day'' to share with their friends.}
% \swelker how to rank these events by rarity and uncommonness?

% \vikas{How many places in Places365 and how many objects in OpenImages?} 365 and 600 respectively -- added above 

\textbf{Egocentric Image Summary Results.} On egocentric images, we show several qualitative examples of summaries generated by our system in Fig. \ref{fig:results}, and compare them to results from a state-of-the-art image captioning model, ClipCap \cite{mokady2021clipcap}.
While state-of-the-art captioning models can perform reasonably over several of the images, we find that our system generally produces more relevant captions for a larger portion of the egocentric examples.
Image captioning models are biased based on the datasets they are trained on, and have shown to perform poorly on egocentric images \cite{agarwal2020egoshots}, which aligns with our observations.
Relatively less research has been carried out specifically on egocentric image captioning \cite{fan2018deepdiary}. SMs can nevertheless produce reasonable captions without additional training on domain-specific data.

{\textbf{\textit{3.3-B. Adding Audio into Single-moment Summaries.}}}
In addition to using visual perceptual inputs, we may use a Socratic approach which engages perceptual inputs from audio as well, via an ALM (audio language model).  Our example egocentric perception system uses Wav2CLIP \cite{wu2021wav2clip} as the ALM.  Wav2CLIP is trained on 5-second audio clips from the VGGSound dataset \cite{chen2020vggsound}, and is trained in a contrastive manner by aligning its audio encoder to the visual CLIP embeddings from video.

\begin{wrapfigure}{r}{0.4\textwidth}
  \vspace{-1.5em}
  \captionsetup{type=figure}
  \includegraphics[width=\linewidth]{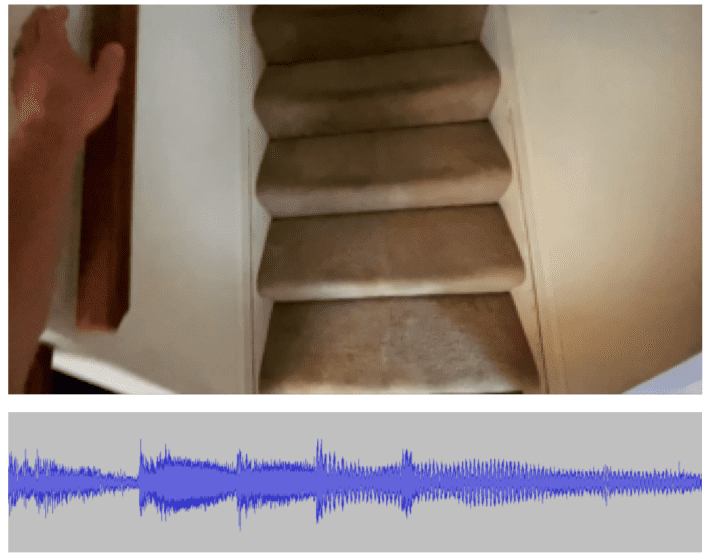}
  \vspace{-1.5em}
  \captionof{figure}{\small Example frame and corresponding (centered) 5-second audio clip which provide the driving example for Sec.~\ref{subsec:linguistic-world-state}-B, \ie adding in ALMs into Socratic dialogue to improve single-moment summarization.  Note that this waveform mostly represents the background piano music, but the system is still able to rank correctly that footsteps as the highest sounds relative to others in the LM-suggested candidate set.}
  \vspace{-7.0em}
  \label{fig:footsteps}
  %https://docs.google.com/drawings/d/1o-TRlbwJiwIJFsCUCLvMJw_jQ_jvvqScevXsJXR_TZo/edit
\end{wrapfigure}

Incorporating an ALM like Wav2CLIP into our Socratic framework can provide an additional modality with which to perform zero-shot cross-modal reasoning, and this may help further improve inference beyond the vision-language-only case.  Fig.~\ref{fig:footsteps} displays a driving example for which a visual-only summarization produced the less-than-desirable summary: ``I am climbing a staircase, and I may see a hamster or human leg'' with the incorrect propogation of the false detection of a hamster and human leg.

To perform audio-aided single-moment summarization, we first run image-based summarization as described previously, but we then prompt the LM to suggest sounds that it may hear, given the visual context, via ``$\langle$\greenuline{visual single-image summary}$\rangle${\color{Gray}. 5 Possible Sounds:}''. For the example in Fig.~\ref{fig:footsteps} an example prompt, which has already gone through multiple rounds of Socratic dialogue to be generated, together with completion by the LM is:

\noindent\fbox{%
    \parbox{0.97\linewidth}{%
        \small{\texttt{\color{Gray}Places: {\color{LimeGreen}staircase}. Objects: {\color{LimeGreen}stairs}, {\color{LimeGreen}animal}, {\color{LimeGreen}mammal}, {\color{LimeGreen}hamster}, {\color{LimeGreen}human leg}.  Activities: {\greenuline{climbing}}.  5 Possible Sounds:} {\texttt{\textbf{\color{RoyalBlue} footsteps, creaking stairs, someone calling your name, a dog barking, a centipede crawling.}}}}
    }%
}

These auditory entities expressed in language can then be ranked by the ALM. In this moment of the video, the sound of footsteps can be faintly heard in the background, and in this case the ALM provides a correct detection of ranking {\purpleuline{footsteps}} as the most likely sound.  This ranking can then be incorporated into a prompt for the LM to provide the single-image summary, for example:

\noindent\fbox{%
    \parbox{0.97\linewidth}{%
        \small{\texttt{\color{Gray}I am in a: {\color{LimeGreen}\{place\}}. I see a: {\color{LimeGreen}\{object1\}}, {\color{LimeGreen}\{object2\}}, {\color{LimeGreen}\{object3\}}, {\color{LimeGreen}\{object4\}}, {\color{LimeGreen}\{object5\}}.  I think I hear {\purpleuline{\{sound1\}}} I am: {\greenuline{\{activity\}}}.  Summary: I am most likely}}
    }%
}

As above, incorporating ``{\color{Gray}I hear {\purpleuline{footsteps}}}'' into the summary and prompting this to the LM provides the completion: ``\textbf{\color{RoyalBlue}climbing a staircase, and I may hear footsteps.}''  In this case, this summary result is preferable to the mentioned single-image caption without sound.

While this example demonstrates in a certain case the utility of audio-informed summaries, overall in egocentric video, with a variety of background noise, we find that Wav2CLIP can provide reasonable detections for certain language-represented auditory entities such as `baby babbling' and entities to do with `running water', but do not provide as robust detections as CLIP.  Also, while there are many advantages to the specific Wav2CLIP approach, including its use of the CLIP embedding space, a major downside is that the training process is ``blind'' to hearing things that cannot be seen.  Accordingly, for the rest of demonstrations shown, we simply build world-state history from VLM-LM interactions alone.  We expect however that with further attention to model approaches, and scaling of audio-language datasets, approaches like Wav2CLIP will increase in robustness.  We also show an additional application (Sec.~\ref{subsec:open-ended-qa}) of audio, for audio retrieval. In that case, only a single auditory search entity is required in order to enable a useful application, and so it can be easier to verify that it is a sufficiently robustly-detected entity.

% \swelker im not sure i fully buy the benefits of including the sounds here. I could imagine if it was speech it could be made a stronger case, but it seems unnecessary for the idea to work, generally its cool that the framework enables these multimodal sensory inputs though.

{\textbf{\textit{3.3-C. Compiling a Language-Based World-State History}}}

Our system compiles the image summaries from each key video frame into a language-based world-state history. Since the total number of frames in the video may be large, compiling a summary for every individual frame would create text that is too large (too many tokens) to be processed directly by an LM as context for Q\&A. Accordingly in this work, we propose solutions that sparsify and/or condense language-based world-state histories (\eg via search-based methods) into practically usable context sizes for reasoning.  In particular, we explore two methods of identifying ``key moments'' in videos for summarization: (i) uniform sampling over time, and (ii) video search (image or audio retrieval) for on-the-fly compilation of context.

The first method, uniform sampling, is straightforward and compiles a world-state history from Socratic summaries of video frames sampled at fixed time intervals. This can also be condensed hierarchically using recursive linguistic summarization \cite{wu2021recursively}, to fit even dense sampling into usable LM-context sizes. However, while broadly indiscriminate, uniform sampling may not have sufficient temporal resolution to capture important spontaneous events in the video (such as adding salt to the pot while cooking soup in the kitchen).

Hence the second method, identifying key moments with video search, uses a VLM or ALM to search for entities most relevant to the question, which can more precisely index the frames in which the subject appears. Specifically, our instantiation of SMs for this component parses a natural language question with an LM into several search entities to be used to find key frames in the video. For example, the question ``{\color{magenta}did I drink coffee today?}'' yields a search entity ``{\textbf{\color{RoyalBlue}drink coffee}}'' that is then used with language-conditioned video search to index the most relevant $n$ key frames of ``{\greenuline{drink coffee}}'' in the video.  The LM categorizes the search, which can be image-based (VLMs) or audio-based (ALMs), \eg for language-conditioned auditory recall questions (\cite{oncescu2021audio}) like ``{\color{magenta}why was my wife laughing today?}'' . While search-based indexing of key moments can be useful for finding spontaneous events, this method for generating context can also provide disadvantages for downstream Q\&A if the answer to the question depends on events that are not directly related to the search subject. For example, ``{\color{magenta}why was I chopping wood today?}'' returns key frames related to ``{\greenuline{chopping wood}}'', but does not return the key frames after the event related to making a campfire. On the other hand, if uniform sampling is employed and the campfire events are captured by the summary, then the LM can successfully return the answer ``{\textbf{\color{RoyalBlue}I was making a campfire.}}'' Choosing which method to use for compiling the language-based world-state history may depend on the application.

\textbf{Language-based World-state History Results.} Fig.~\ref{fig:results}, middle, shows results generated by our system. The specific event log shown in Fig.~\ref{fig:results} has been trimmed down for space considerations, but is representative of the type of event logs that may be generated without manual curation.  These event logs are used as context to enable LM open-ended reasoning on video, as demonstrated in the next section.

\begin{figure*}[t]
  \centering
  \vspace{0.5em}
  \includegraphics[width=0.99\linewidth]{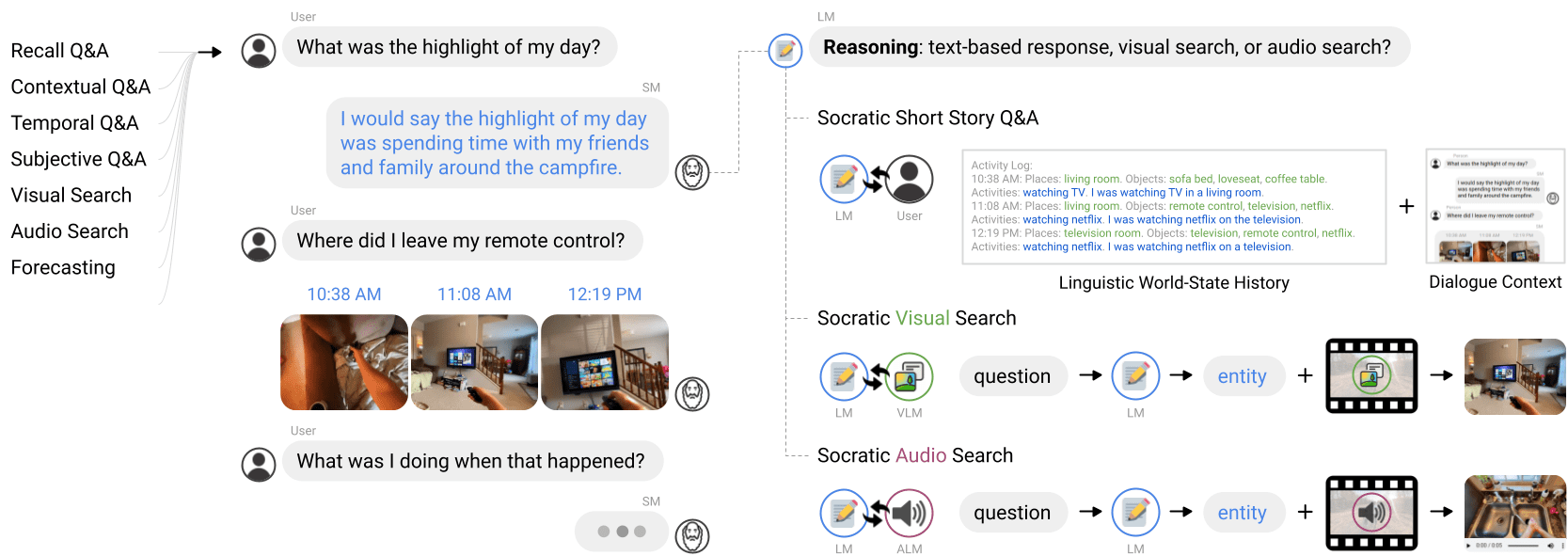}  
  \caption{\small{SMs can interface with the user through dialogue and perform a variety of tasks (formulated as Q\&A) with egocentric video: sorting reasoning questions by their output modalities \eg text-base responses, images from visual search, video snippets from audio search. Depending on the modality, each question can pass through a different sequence of Socratic interactions between the LM, VLM, and ALM.
  }}
  \label{fig:qa-system}
\end{figure*}

\subsection{Open-Ended Reasoning on Egocentric Video}\label{subsec:open-ended-qa}

In this section we describe a few examples of how the Socratic Models framework can be used to perform open-ended multimodal-informed completion of text prompts, conditioned on egocentric video (examples in Fig.~\ref{fig:teaser}). %We first describe how the system can provide answers to a diverse set of questions that are essentially video understanding.
There are of course limitations to what they can provide, but our demonstrated examples suggest that we can already today generate compelling answers to open-ended reasoning tasks, at a scope that is beyond what we are aware is possible today with available methods. Of course, the answers may also inherit undesirable characteristics from the component models, such as an LM that is overconfident even when wrong. It is our hope that our results may help inspire work on preparing even more comprehensive video understanding datasets for the community, to assist further assessment.

Our example system uses a language-based world-state history generated through Socratic multi-model discussion (Sec.~\ref{subsec:linguistic-world-state}), and provides this as context to an LM to enable open-ended reasoning on egocentric videos.
Open-ended text prompts from a user, conditioned on an egocentric video, can yields three types of responses: a text-based response, a visual result, and/or an audio clip.  %These latter two are examples of {\em{video search}} and can for example be addressed by Socratic multimodal analysis of the full, dense video. 
These latter two provide examples that open up the capabilities of the system to respond not only with text-based responses, but also respond with video snippets themselves, which may be a higher-bandwidth way to respond to user requests ({\em{``a picture is worth a thousand words''}}).  %Then we show how with simple zero-shot LM-based classification of a User question, the system can alternatively be composed to provide {\em{either}} text-based responses, or dense image and/or audio retrieval.
The specific composition of our system is of course just one example -- overall, the modularity of the Socratic approach makes it easy to compose together foundation models, zero-shot, in a variety of ways to provide a spectrum of multimodal reasoning capabilities.

The demonstrated tasks include (i) summarization, (ii) open-ended Q\&A, (iii) forecasting, (iv) corrections, and (v) video search for either visual or audio cues. These tasks have predominantly been studied in isolation in the research community -- but our example results with SMs suggest they can be subsumed under the same unified language-based system for multimodal reasoning. 
% {\em{\textbf{Descriptions and results}}} for each of (i)-(v) are shown below.

%where the answers may include diverse video understanding questions that require synthesis of full-day context, or may instead retrieve specific moments (video or audio search).

%To demonstrate a variety of capabilities, we discuss how the system can provide a variety of answers to open-ended Q\&A, 

%While image and audio retrieval have been studied separately, we show how they can be formulated as Socratic multi-model discussions, and additionally show their integration into a system that can selectively detect when such questions are asked.

%{\textbf{\textit{2.4-A. Text-based Responses: Video Understanding as Reading Comprehension.}}} By providing our language-based world-state history as context, we can directly use an LM for downstream text-based response tasks including (i) summarization, (ii) diverse Q\&A, and (iii) forecasting.  These tasks have predominantly been studied in isolation -- but these example results with SMs suggest they can be subsumed under the same unified language-based system for visual reasoning. 

\textbf{(i) Summarization} can be implemented by prompting an LM to complete the excerpt ``{\{\greenuline{world-state history}\}} {\color{Gray} Summary of my day:}'' to which it can respond with outputs like ``{\textbf{\color{RoyalBlue}I slept in a bed, made coffee, watched TV, did laundry, received a package, bench pressed, showered, ate a sandwich, worked on a computer, and drank wine.}}'' Since the language-based world-state history is constructed with summaries of visual content, it carries contextual information that can be complementary to what is found in closed captions (\eg speech and dialogue). Summarizing egocentric videos enables a number of applications, including augmenting human memory to recall events, or life-logging of daily activities for caregiver assistance. Our system draws similarity to early work in the area involving text-based summarization and identifying key frames (see \cite{barbieri2003video} for an early survey and \cite{del2016summarization,apostolidis2021video} for more recent surveys).
%Summarizing egocentric videos has many applications:  to instructional video navigation and remote assistance ().
%\michael{Earlier works on egocentric video summarization focused on...}

%{\textbf{\textit{3.4-A. Open-Ended Q\&A on Egocentric Video}}}

\textbf{(ii) Open-ended Q\&A} can be implemented by prompting the LM to complete the template: ``{\{\greenuline{world-state history}\}} {\color{Gray}Q:} {\color{magenta}\{question\}} {\color{Gray}A:}''. We find that LMs such as GPT-3 can generate surprisingly meaningful results to binary yes or no questions, contextual reasoning questions, as well as temporal reasoning questions. As in \cite{yang2021empirical} we can further prompt the LM to \textit{explain the answer} by adding ``{\color{Gray}This is because:}''. We find that the accuracy of the answers and explanations remain largely conditioned on whether the necessary information can be found within the world-state history. This suggests that the quality of the language-based reconstructions of the videos (\eg via key frame sampling and captioning in this work) is central to the approach. %\johnny{are the examples of free form captured in Fig 4 and Fig 5? if so, would be good to reference}

We show qualitative examples of free-form question answering using our SM system on egocentric video in Fig.~\ref{fig:results}, bottom, Fig.~\ref{fig:system}, and Fig.~\ref{fig:qa-system} generated using a first-person POV video\footnote{\small Examples on \href{https://youtu.be/-UXKmqBPk1w}{https://youtu.be/-UXKmqBPk1w} used with permission from Cody Wanner.} as input. 

\textbf{\textit{Recall Questions.}} SMs can perform simple retrieval of events.
For example, ``{\color{magenta}did I eat dinner today?}'', yields a response ``\textbf{{\color{RoyalBlue}yes I ate dinner today.}}'' along with an explanation ``\textbf{{\color{RoyalBlue}I was seen eating a sandwich in a kitchen at 5:27 PM.}}'' which points to the key frame that was captioned with the sandwich in hand.
Another example that involves contextual reasoning to recall events is ``{\color{magenta}what was I doing outdoors?}'' to which the system responds ``{\textbf{\color{RoyalBlue}I was chopping wood in a yard.}}'' 
Likewise, if the entities described in the question do not appear in the world-state history, such as ``{\color{magenta}did I drive today?}'' the system can respond with a negative answer: ``\textbf{{\color{RoyalBlue}no, I did not drive today.}}'' with an explanation ``\textbf{{\color{RoyalBlue}I was at home all day.}}''
This capability expands beyond standard video search, which might only return nearest neighbor video frames, without a natural language response (or a negative response).

The performance of recalling events largely depends on the relevance of the language-based world-state history to the question. %, constructed by summarizing key frames.
% Thus different key frame sampling strategies can influence output answers.
We find that recall-type questions work best with world-state history logs that are compiled by using search-based key frame indexing (see Sec. 3.3-B).
The system can still return negative responses, since the captioning of the key frames are not influenced by the question.

\textbf{\textit{Temporal Reasoning.}} SMs can answer questions related to time by appending timestamps to each key moment in the world-state history.
By associating image summaries to times of the day, this allows answering questions that time index various activities.
For example {\color{magenta}``when did I last drink coffee?}'' can return the last time drinking coffee was mentioned in the log, with a full response ``\textbf{{\color{RoyalBlue}I last drank coffee at 10:17 AM}}'' and an explanation ``\textbf{{\color{RoyalBlue}I was making coffee in the kitchen.}}''
The system can also count events, for example when asked ``{\color{magenta}how many times did I receive a package today?}'', the system will respond appropriately ``\textbf{{\color{RoyalBlue}I received a package once today.}}'' with an explanation ``\textbf{{\color{RoyalBlue}I was receiving a package at 3:24 PM}}''.
We find that a common failure mode for these types of questions is that the system tends to over-count, especially as a reaction to false positive VLM detection results that get surfaced into the world-state history. For example, asking ``{\color{magenta}who did I interact with?}'' would yield ``\textbf{{\color{RoyalBlue}woman, hamster}}'' where hamster was a false positive prediction from CLIP.
These issues become more prominent with search-based key frame sampling, as a byproduct of an inability to distinguish neighboring local argmaxes of the same event from each other.

\textbf{\textit{Cause and Effect Reasoning.}} SMs can answer questions about cause and effect relationships between events, conditioned on that all the events appear in the world-state history.
For example, when asked % \pete{suggest to change this to ``why did I go to the porch'' because basically the only reason to chop wood it to make a fire, but there may be many reasons to go to the front door.} 
``{\color{magenta}why did I go to the front porch today?}'' the system would respond ``\textbf{{\color{RoyalBlue}I went to the front porch today to receive a package.}}'' and an explanation ``\textbf{{\color{RoyalBlue}I saw on the porch a package and knew that I was expecting it.}}'' These types of questions are exciting because they suggest opportunities for prompting logical deduction of events. However, since information about both the cause and the effect needs to be in the world-state history, the quality of results remains highly dependent on the key frame sampling strategy used to compile it (Sec. 3.3-B). Uniform gives an unbiased account of events, and is currently the best variant for this form of reasoning. More targeted construction of the world-state history with search based key frames can sometimes miss frames that capture the answer to the question.

\textbf{\textit{Subjective Reasoning.}} SMs can also answer more subjective questions, such as ``{\color{magenta}was I happy today?}'' or ``{\color{magenta}what was my favorite drink today?}''. Without additional context, these questions rely on biases from the LM's dataset -- which could have negative consequences, and should be managed carefully with additional mechanisms for safety and groundedness \cite{thoppilan2022lamda}. The full personalization of these subjective questions are likely to be conditioned on whether a better context can be constructed of prior user behaviors related to the question.

\textbf{(iii) Forecasting} of future events can be formulated as language-based world-state completion. Our system prompts the LM to complete the rest of an input event log. Timestamps of predictions can be preemptively specified depending on application needs. The completion results are generative, and more broad than binary event classification (\eg \cite{lei2020more}). Example completion (also shown in Fig. \ref{fig:teaser}):

\noindent\fbox{%
    \parbox{0.97\linewidth}{%
        \small\texttt{{\color{Gray}1:46 PM: I am eating a sandwich in a kitchen.\\
2:18 PM: I am checking time and working on a laptop in a clean room.\\
2:49 PM: I am buying produce from a grocery store or market.\\
3:21 PM:} {\textbf{\color{RoyalBlue} I am driving a car.\\
4:03 PM: I am in a park and see a playground.\\
4:35 PM: I am in a home and see a television.}}}
    }%
}

Few-shot prompting the LM with additional examples of prior event logs most similar to the current one is likely to improve the accuracy of the completion results. Without additional context, these results are again biased towards typical schedules seen by the LM across Internet-scale data.

To a certain extent, this forecasting capability extends and generalizes the traditional topic of activity forecasting in computer vision. In the research community, activity forecasting has been often formulated as an extension of action classification, tracking, or feature generation: Given a sequence of image frames, they directly predict a few categorized actions \cite{ryoo2011human,hoai2014max,rhinehart2017first}, human locations \cite{kitani2012activity}, or image features \cite{vondrick2016anticipating} to be observed in the future frames. In contrast, Socratic Models with LMs enables generating more semantically interpretable descriptions of future events, conditioned on multimodal information.

%\michael{Discuss forecasting in general in computer vision. Forecasting with egocentric videos:} The objective of activity forecasting is to infer humans/their actions/objects to appear in the future scene, given the current context. In Vision, this has been approached either as an extension of action classification, tracking, or pixel generation. That is, given a sequence of image frames, the approach was to directly predict a few categorized actions \cite{}, human locations \cite{}, or image frames \cite{} to be observed in the future.

% We show an example of SMs forecasting the event log in Fig. \ref{fig:teaser}, which relies on the timestamps to forward predict future activities. 

\textbf{(iv) Corrections.} SMs can be prompted to incorporate {\color{magenta}human feedback} in the loop as well, which could be useful for interactive language-based systems. For example, given image captions generated from an VLM and LM:

\noindent\fbox{%
    \parbox{0.97\linewidth}{%
        \small\texttt{{\color{Gray}Context: Where am I? outdoor cabin, campsite, outdoor inn. What do I see? fire, marshmallow, fire iron, hearth, fireside, camp chair. What am I doing? Commonsense suggests: roasting marshmallows, sitting around the fire, chatting. Most likely: sitting around the fire. \\Original Summary:} {\color{RoyalBlue}I am camping and enjoying the company of my friends around the fire.} \\{\color{Gray}Corrections:} {\color{magenta}It was actually my family, not friends, sitting around the fire.} \\{\color{Gray}Corrected Summary:} {\textbf{\color{RoyalBlue}I am camping with my family and enjoying the company of them around the fire.}}}
    }%
}

\begin{figure}[t]
  \centering
  %\vspace{2.5em}
  \captionsetup{type=figure}
  \includegraphics[width=0.7\linewidth]{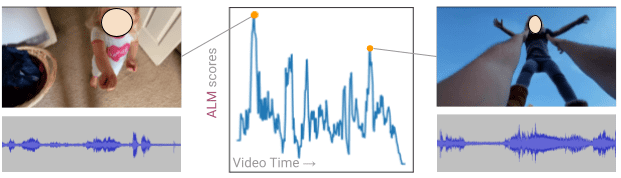}
  \vspace{-1.0em}
  \captionof{figure}{\small Example zero-shot language-prompted auditory retrieval (shown: top 2 results) in response to ``{\color{magenta}what did my daughter's laugh sound like today?}'', for which an LM identifies the audio search query of ``{\purpleuline{daughter's laugh}}'', and an ALM (Wav2CLIP) is used for audio retrieval.  The top (left) retrieval is only partially correct, returning a video clip involving the daughter but not laughter.  The second (right) retrieval is correct, from a moment of playing (getting tossed into the air).  Faces obscured for privacy.
  % https://docs.google.com/drawings/d/1to27BwBlMe1tQi3YCq1P21EaTy0uKZJeLOfncV0KsH0/edit
  }
  \label{fig:laughter-recall}
  \vspace{-2.0em}
\end{figure}

\textbf{(v) Video Search: Image or Audio Retrieval.} Our SM system can also return additional modalities (images, audio) as answers to questions, by simply few-shot prompting the LM to classify a target modality based on the input question. For example, ``{\color{magenta}where did I leave my remote control}'' can map to image search using VLM features for ``{\greenuline{remote control}}'' while ``{\color{magenta}what did my daughter's laugh sound like today?}'' % johnny{modified query to request an audio result - although variant is a quite an emotionally sad, though empathisable, query! this query alone could be a commercial for this system}
can map to natural-langauge-queried audio search (\cite{oncescu2021audio}) using ALM features for ``{\purpleuline{daughter's laugh}}'' (Fig.~\ref{fig:laughter-recall}). 
This can be useful for some applications (\eg AR) in which the user may find the retrieved modality to be more useful than a natural language response. 
Our approach for this uses an LM to parse a search entity from the question to index key video frames. This is done with several few-shot examples provided as context. For example, the question ``{\color{magenta}when did I last wash my hands?}'' yields a search entity ``{\textbf{\color{RoyalBlue}wash my hands}}'' that is then used with video search to index the most relevant $n$ key frames of ``{\greenuline{wash my hands}}'' in the video.
Specifically, our system runs video search by ranking matching CLIP or Wav2CLIP features of the entity text against all video frames, and returning the top $n$ local maximums.
For each frame, the features can either be image features or audio features (\eg from the surrounding 5 seconds with Wav2CLIP) -- where the LM few-shot categorizes which domain to use for any given question. This can be thought of as calling different subprograms for hierarchical search.

\textbf{Limitations.} Overall, our results suggest that SMs are capable of generating meaningful outputs for various egocentric perception tasks via visual contextual reasoning -- but its limitations also suggest areas for future work. For example, a primary bottleneck in the Q\&A system is that it relies on the richness (\ie recall) and quality (\ie precision) of the event log. This likely could be improved with better image and audio detectors or captioning systems \cite{gu2021open}.  Also, we find that the used Wav2CLIP may provide satisfactory results for certain categories in audio retrieval, but we currently do not involve it in generating the event log, since its robustness and range of open-language detection is not at the same level as CLIP.  This seems addressable with further approaches and scaling of datasets in the audio-language domain.

Additionally, accurate response to cause and effect reasoning questions also require relevant key moments to be reflected in the event log -- which points to open ended questions on how to achieve better key frame sampling (beyond the simple baselines that we have demonstrated). Finally, the dialogue between the different models are fairly structured with manually engineered prompts. It may be interesting to investigate more autonomous means of achieving language-based closed loop discussions between the models until a commonsense consensus is reached.

\section{Scaling Up Socratic Video Search} \label{subsec:scaling-video-search}

The search algorithms of the SMs, which may be used both for compiling world-state history (Sec.~\ref{subsec:linguistic-world-state}-C) and for video search retrieval (Sec.~\ref{subsec:open-ended-qa}) rely on the matching procedure conducted in the corresponding latent space (\eg VLM features of the text snippet against these of the video frames). This can be abstracted as dot-product-maximization key search in the given key-dataset. In practice, if the key-dataset is large (\eg long videos) a naive linear search is prohibitively expensive. We propose several solutions to this problem. 

\vspace{-3mm}
\paragraph{MIP-Search.} The first observation is that several data pre-processing techniques applied in the so-called \textit{maximum inner product} (MIP) search can be directly used to reorganize the keys (\eg latent representations of video frames) to provide sub-linear querying mechanism for the incoming text snippet (see: \cite{mip-search}). Those include pruning and various indexing techniques, such as LSH-hashing \cite{asymmetric-hashing}. In the hashing approach, a collection of hash-tables, indexed by the binarized representations of the hashes is stored with different entries of the hash table corresponding to the subsets of keys producing a particular hash. There are several cheap ways of computing such hashes, \eg \textit{signed random projection} (those in principle linearize the angular distance, but every MIP task can be translated to the minimum angular distance search problem). The querying is then conducted by searching for the most similar hash-entries in the hash-tables and then performing linear search only on the subsets of keys corresponding to these entries to obtain final ranking.  
\vspace{-3mm}
\paragraph{Associative Memories.} The above approach provides sub-linear querying mechanism, but does not address the space complexity problem.
In the scenario of strict memory requirements, we propose to leverage recently introduced techniques on linear attention \cite{performers} combined with \textit{modern continuous associative memory} (MCAM) models \cite{modern-hopfield}. MCAM models are de facto differentiable dictionaries (with provable few-shot retrieval) that can be thought of as energy-based models using negated exponentiated latent-representations-dot-product energy for the \textit{exponential} storage capacity. A naive computation of such an energy still requires explicitly keeping all the patterns (which is exactly what we want to avoid), but this can be bypassed by applying the linearization of that energy (which effectively is just the negated sum of the softmax kernel values) with the FAVOR+ mechanism used in linear-attention Transformers, called \textit{Performers} \cite{performers}. This modification has several advantages: (1) it makes the size of the dictionary completely independent from the number of the implicitly stored patterns; the size now scales linearly with the number of random features used for energy linearization, (2) it provides a \textit{constant-time} querying mechanism at the price of compressing all the patterns (and thus losing some information). 
\vspace{-3mm}
\paragraph{Random Feature Trees.} The other approach, that combined the ideas from both MIP-search and linear attention systems, leverages the so-called \textit{random feature tree} (RFT) data structure \cite{rawat}. This approach relaxes the MIP-search to sampling from the linearized softmax distribution via FAVOR+ \cite{hrfs}. Sampling from such a linearized distribution can be done in time logarithmic in the number of samples via RFT which is a balanced tree with leaves corresponding to latent representations of video frames and nodes encoding representations of the subsets of keys (\eg the video frames) defined as sums of the random feature transforms of the keys.

\section{Additional Notes on Robot Experiments}

\begin{figure}[H]
  \centering
  \vspace{-1.5em}
  \captionsetup{type=figure}
  \includegraphics[width=1.0\linewidth]{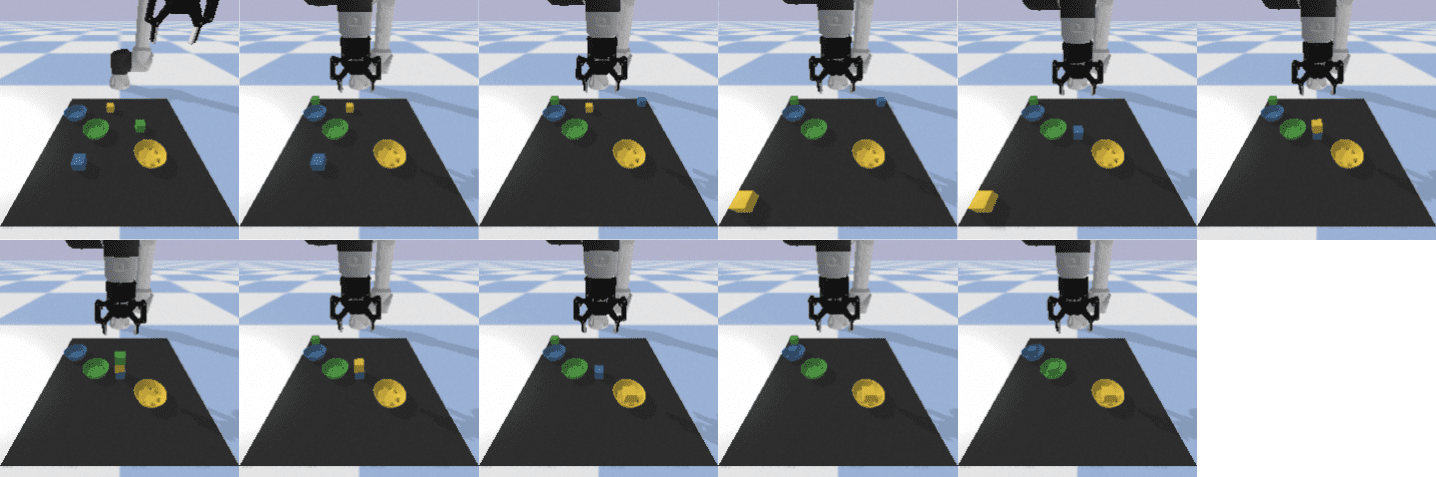}
  \vspace{-1.0em}
  \captionof{figure}{\small Full rollout of the robot environment for the example presented in Sec. 5.3 of the main paper.
  }
  \label{fig:robot-rollout}
  \vspace{-1.5em}
\end{figure}

The SM robot system uses a VLM (open-vocabulary object detection with ViLD \cite{gu2021open}) to describe the objects in the scene, feeds that description as context to a LM as a multi-step planner \cite{saycan2022arxiv,huang2022language}, that then takes as input a natural language instruction and generates the individual steps to be passed to a pretrained language-conditioned robot policy, for which we specifically use a CLIP-conditioned \cite{shridhar2022cliport} No-Transport baseline from Zeng et al. \cite{zeng2020transporter} (inspired by CLIPort \cite{shridhar2022cliport} for open vocabulary pick-and-place). The full prompt used as context to the LM for multi-step planning is:

\noindent\fbox{%
    \parbox{0.97\linewidth}{%
        \scriptsize\texttt{{\color{Gray}
objects = ["cyan block", "yellow block", "brown block", "green bowl"] \\
\# move all the blocks to the top left corner. \\
Step 1. robot.pick\_and\_place("brown block", "top left corner") \\
Step 2. robot.pick\_and\_place("cyan block", "top left corner") \\
Step 3. robot.pick\_and\_place("yellow block", "top left corner") \\
\# put the yellow one the green thing. \\
Step 1. robot.pick\_and\_place("yellow block", "green bowl") \\
\# undo that. \\
Step 1. robot.pick\_and\_place("yellow block", "top left corner") \\
objects = ["pink block", "gray block", "orange block"] \\
\# move the pinkish colored block on the bottom side. \\
Step 1. robot.pick\_and\_place("pink block", "bottom side") \\
objects = ["orange block", "purple bowl", "cyan block", "brown bowl", "pink block"] \\
\# stack the blocks. \\
Step 1. robot.pick\_and\_place("pink block", "orange block") \\
Step 2. robot.pick\_and\_place("cyan block", "pink block") \\
\# unstack that. \\
Step 1. robot.pick\_and\_place("cyan block", "bottom left") \\
Step 2. robot.pick\_and\_place("pink block", "left side") \\
objects = ["red block", "brown block", "purple bowl", "gray bowl", "brown bowl", "pink block", "purple block"] \\
\# group the brown objects together. \\
Step 1. robot.pick\_and\_place("brown block", "brown bowl") \\
objects = ["orange bowl", "red block", "orange block", "red bowl", "purple bowl", "purple block"] \\
\# sort all the blocks into their matching color bowls. \\
Step 1. robot.pick\_and\_place("orange block", "orange bowl") \\
Step 2. robot.pick\_and\_place("red block", "red bowl") \\
Step 3. robot.pick\_and\_place("purple block", "purple bowl")
}}
    }%
}

\begin{figure}[H]
  \centering
  %\vspace{2.5em}
  \captionsetup{type=figure}
  \includegraphics[width=1.0\linewidth]{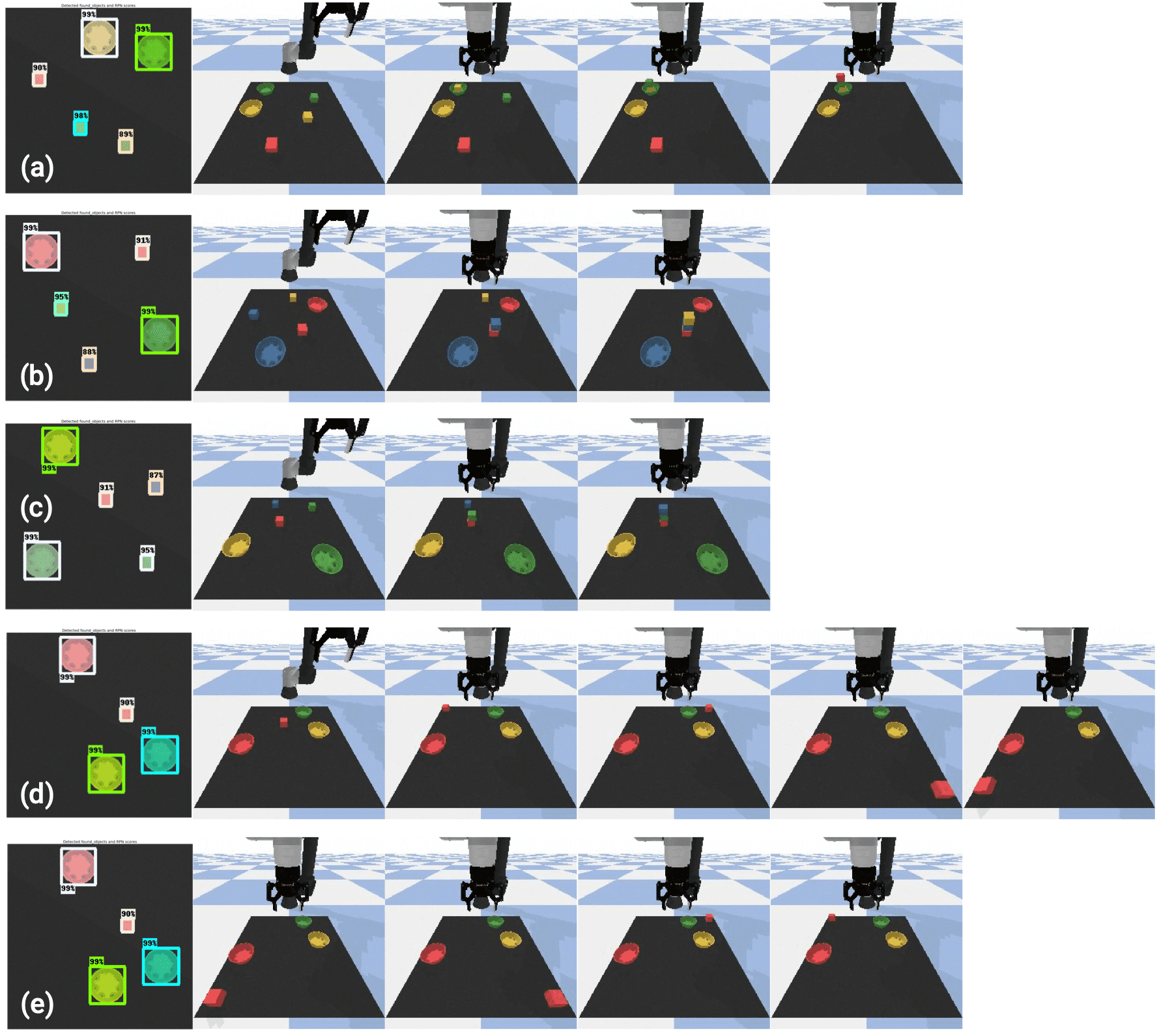}
%   \vspace{-1.0em}

  \noindent\fbox{%
    \parbox{0.98\linewidth}{%
        \scriptsize\texttt{{\color{Gray}\\
(a)\\
objects = [{\color{LimeGreen}'green bowl', 'yellow bowl', 'yellow block', 'green block', 'red block'}]\\
\# {\color{magenta}Put all the blocks in the green bowl.}\\
Step 1. {\color{RoyalBlue}robot.pick\_and\_place('yellow block', 'green bowl')}\\
Step 2. {\color{RoyalBlue}robot.pick\_and\_place('green block', 'green bowl')}\\
Step 3. {\color{RoyalBlue}robot.pick\_and\_place('red block', 'green bowl')}\\
\\
(b)\\
objects = [{\color{LimeGreen}'red bowl', 'blue bowl', 'yellow block', 'red block', 'blue block'}]\\
\# {\color{magenta}Stack all the blocks.}\\
Step 1. {\color{RoyalBlue}robot.pick\_and\_place('blue block', 'red block')}\\
Step 2. {\color{RoyalBlue}robot.pick\_and\_place('yellow block', 'blue block')}\\
\\
(c)\\
objects = [{\color{LimeGreen}'green bowl', 'yellow bowl', 'green block', 'red block', 'blue block'}]\\
\# {\color{magenta}Stack all the blocks.}\\
Step 1. {\color{RoyalBlue}robot.pick\_and\_place('green block', 'red block')}\\
Step 2. {\color{RoyalBlue}robot.pick\_and\_place('blue block', 'green block')}\\
\\
% (d)\\
% objects = [{\color{LimeGreen}'red bowl', 'yellow block', 'green block', 'red block', 'blue block'}]\\
% \# {\color{magenta}Put all the blocks in different corners.}\\
% Step 1. {\color{RoyalBlue}robot.pick\_and\_place('yellow block', 'top left corner')}\\
% Step 2. {\color{RoyalBlue}robot.pick\_and\_place('green block', 'top right corner')}\\
% Step 3. {\color{RoyalBlue}robot.pick\_and\_place('red block', 'bottom left corner')}\\
% Step 4. {\color{RoyalBlue}robot.pick\_and\_place('blue block', 'bottom right corner')}\\
% \\
(d)\\
objects = [{\color{LimeGreen}'green bowl', 'red bowl', 'yellow bowl', 'red block'}]\\
\# {\color{magenta}Clockwise, move the block through all the corners.}\\
Step 1. {\color{RoyalBlue}robot.pick\_and\_place('red block', 'top left corner')}\\
Step 2. {\color{RoyalBlue}robot.pick\_and\_place('red block', 'top right corner')}\\
Step 3. {\color{RoyalBlue}robot.pick\_and\_place('red block', 'bottom right corner')}\\
Step 4. {\color{RoyalBlue}robot.pick\_and\_place('red block', 'bottom left corner')}\\
\\(e)\\
\# {\color{magenta}Now, move it back counterclockwise.}\\
Step 5. {\color{RoyalBlue}robot.pick\_and\_place('red block', 'bottom left corner')}\\
Step 6. {\color{RoyalBlue}robot.pick\_and\_place('red block', 'bottom right corner')}\\
Step 7. {\color{RoyalBlue}robot.pick\_and\_place('red block', 'top right corner')}\\
Step 8. {\color{RoyalBlue}robot.pick\_and\_place('red block', 'top left corner')}
}}
    }%
}

  \captionof{figure}{\small Additional examples of multi-step tasks that the SM robot system can perform out-of-the-box. 
  }
  \label{fig:robot-appendix}
  \vspace{-1.0em}
\end{figure}

Fig. \ref{fig:robot-rollout} depicts a full rollout of the example in Sec. 5.3 in the main paper, which involves human dialogue. Fig. \ref{fig:robot-appendix} shows additional examples of multi-step tasks that the system can perform out-of-the-box with zero-shot SMs.
The system is able to reason over order and nuanced language (clockwise vs. counterclockwise) as well as respond to different objects being detected (the stacking task with varied block colors).
Note that the LM is few-shot prompted to generate pseudo code ``{\color{RoyalBlue}robot.pick\_and\_place("A", "B")}'' which calls a function to return a fixed template sentence ``Pick the A and place it on the B.'' subsequently fed as input to the language-conditioned robot policy. While we can prompt the LM to directly produce the template sentences as opposed to code, we find that the LM can sometimes generate phrases or prepositions that are beyond the training set of the language-conditioned policy. We observe that the policies are more likely to return correct actions when the templates can be engineered to be more similar to the phrases seen within the policy's training data.
We also found ViLD and CLIP to be brittle in this scene, as the scene is simulated and the objects are not natural. High-performance in this setting requires a good view angle (an overhead camera), filtered colors (red, green, yellow, and blue), and tuned names (we referred to the blocks as ``boxes'' and the bowls as ``circles'' to account for the overhead view).
Without these changes, the system still is able to complete many tasks, but less consistently.  We expect off-the-shelf VLMs of the future to be more robust than those currently available.

\section{Socratic Deductive Reasoning}

%\vincent{Mention the 'chain of thoughts prompting' paper and implications for using LMs for reasoning}
In the context of egocentric perception, we find that formulating video Q\&A as reading comprehension in SMs directly leverages the extent to which large LMs are capable of logical reasoning by connecting commonsense relationships with knowledge learned from Internet-scale data. For example, the system returns the following answer when presented with the world-state history log:
%\vikas{suggesting using more prominent color below, and split into lines.}

\noindent\fbox{%
    \parbox{0.97\linewidth}{%
        \small\texttt{{\color{Gray}
8:00 AM: went to grocery store to buy orange juice, chocolate, and bread.\\
8:15 AM: I went to gas station to fill up the vehicle tank.\\
8:30 AM: drove back home and left the groceries in the kitchen.\\
8:45 AM: started cooking eggs in the pan.\\
9:00 AM: the dog went into the kitchen.\\
9:15 AM: took the dog out for a walk.\\
9:30 AM: the dog is sick.\\
Q: Why is the dog sick? A:} {\textbf{\color{RoyalBlue}The dog may have eaten something it was not supposed to, such as chocolate.}}}
    }%
}

Arriving at the answer requires bridging multiple connections between observations \eg the dog went into the kitchen, the groceries are still in the kitchen, and the groceries contain chocolate.
Such results offer a glimpse of what might be possible using SMs for deductive reasoning across multiple domains of information, and raises interesting research questions on (i) how to better assemble language-based world-state histories (beyond what is presented in this work) that capture relevant evidence to improve the accuracy of conclusions, and (ii) how to elicit chain of thought prompting \cite{wei2022chain} to decompose multi-step problems into intermediate ones. For example, one promising extension could be prompting the LM with chain of thought sequences to expand on hypotheses:

\noindent\fbox{%
    \parbox{0.97\linewidth}{%
        \small\texttt{{\color{Gray}Q: What are reasons for why I might be chopping wood? A: Reasons might include:} {\textbf{\color{RoyalBlue}needing firewood, wanting to make a statement, or needing the exercise.}}}
    }%
}

to which each hypothesis can be progressively explored by downstream subprograms called at recursively higher resolutions until a conclusion is reached.
These directions suggest pathways towards achieving increasingly meaningful utility and analysis by digital multimodal assistants.

\section{Broader Impact: Energy and Resource Consumption}

Regarding the impact on energy and other resource consumption, this work may help pave a path for new, capable machine learning models to be composed with minimal training resource consumption, provided that large foundational pretrained models are available. This may help provide an answer for how large pretrained models may be retargeted to a wide variety of multimodal applications, without additional considerable compute resources required.
Since SMs help demonstrate how a wide variety of applications may be addressed with fixed (pretrained) models zero-shot, this may also help foster adoption of new machine learning accelerators (\eg fixed analog circuity \cite{reuther2020survey}, optical diffraction \cite{lin2018all}) for inference with substantially lower power consumption and more compact form factors.